\documentclass{article}
\usepackage{iclr2021_conference,times}

\usepackage{amsmath,amsfonts,bm}

\def\eqref#1{equation~\ref{#1}}
\def\1{\bm{1}}

\DeclareMathAlphabet{\mathsfit}{\encodingdefault}{\sfdefault}{m}{sl}
\SetMathAlphabet{\mathsfit}{bold}{\encodingdefault}{\sfdefault}{bx}{n}

\usepackage{times}
\usepackage{latexsym}

\usepackage{url}

\usepackage{verbatim}
\usepackage{microtype}
\usepackage{graphicx}
\usepackage{booktabs} % for professional tables

\usepackage{hyperref}

\usepackage{amsmath, amssymb, amsfonts, amsthm}
\usepackage[utf8]{inputenc} % allow utf-8 input
\usepackage[T1]{fontenc}    % use 8-bit T1 fonts
\usepackage{bbm}
\usepackage{graphicx}
\usepackage{color}
\usepackage{bm}
\usepackage{float}
\usepackage{graphicx}
\usepackage{mathabx}
\usepackage{multirow}
\usepackage{setspace}

\usepackage{booktabs}
\usepackage{nicefrac}       % compact symbols for 1/2, etc.
\usepackage{microtype}      % microtypography
\usepackage{graphicx} % more modern
\usepackage{subfig}
\usepackage{adjustbox}  
\usepackage{wrapfig,lipsum,booktabs}
\usepackage{tikz}
\usetikzlibrary{arrows,shapes,automata,backgrounds,petri}
\usepackage{multirow}% http://ctan.org/pkg/multirow
\usepackage{hhline}% http://ctan.org/pkg/hhline

\usepackage{wrapfig}
\usepackage{makecell}
\usepackage{svg}

\usepackage{multirow}
\usepackage{balance}
\usepackage{amsfonts}
\usepackage{amsmath}
\usepackage{amssymb}
\usepackage{graphicx}
\usepackage{microtype}
\usepackage{xspace}
\usepackage{wrapfig,lipsum}
\usepackage{algorithm}
\usepackage{algorithmicx}
\usepackage{algpseudocode}
\usepackage{makecell}
\usepackage[T1]{fontenc}
\usepackage{soul}
\usepackage{tabularx, booktabs}
\usepackage{trimclip}

\usepackage{hyperref}
\usepackage{todonotes}
\usepackage{caption}

\newcolumntype{Y}{>{\centering\arraybackslash}X}
\newcolumntype{L}{>{\arraybackslash}X}

\algnewcommand\algorithmicinput{\textbf{Input:}}
\algnewcommand\algorithmicoutput{\textbf{Output:}}
\algnewcommand\algorithmicparameter{\textbf{Parameters:}}
\algnewcommand\INPUT{\item[\algorithmicinput]}
\algnewcommand\OUTPUT{\item[\algorithmicoutput]}
\algnewcommand\PARAMETER{\item[\algorithmicparameter]}

\newcommand\ModelName{DeBERTa}
\newcommand\DecoderName{EMD}

\iclrfinalcopy
\title{\ModelName: Decoding-enhanced BERT with Disentangled Attention}
\author{Pengcheng He$^1$, Xiaodong Liu$^2$, Jianfeng Gao$^2$,  Weizhu Chen$^1$ \\
    $^1$ Microsoft Dynamics 365 AI  ~~~~~~~~  $^2$ Microsoft Research \\
  {\tt \{penhe,xiaodl,jfgao,wzchen\}@microsoft.com}
}

\date{December 2019}
\graphicspath{figures/}
\DeclareGraphicsExtensions{.png,.pdf,.svg}

\begin{document}

\maketitle

\begin{abstract}
Recent progress in pre-trained neural language models has significantly improved the performance of many natural language processing (NLP) tasks. In this paper we propose a new model architecture \textbf{DeBERTa}
%, which stands for 
(\textbf{D}ecoding-\textbf{e}nhanced \textbf{BERT} with disentangled \textbf{a}ttention) 
%Different from the BERT\cite{devlin2018bert} or RoBERTa\cite{liu2019roberta}, 
that improves the BERT and RoBERTa models using two novel techniques.
% the language model pre-training of a masked language model. 
%First, instead of using absolute position embedding to model sequence dependency, DeBERTa applies a new relative position encoding method called disentangled attention which thoroughly captures the dependency between positions and content in the sequence. 
The first is the disentangled attention mechanism, where each word is represented using two vectors that encode its content and position, respectively, and the attention weights among words are computed using disentangled matrices on their contents and relative positions, respectively.
Second, 
%inspired by the encoder-decoder attention in original transformer structure designed for seq-to-seq tasks, 
an enhanced mask decoder is used to incorporate absolute positions in the decoding layer to predict the masked tokens in model pre-training.
%an enhanced mask decoder is used to replace the output softmax layer 
%the last layer of Transformer $decoder$ 
%to decode masked tokens for model pre-training. 
% which encourage the self-attention to capture global contextual information instead of looking at itself.
%on top of existing multi-layer transformer encoder, DeBERTa introduces the notion of multi-layer decoding  to better reconstruct masked tokens. 
In addition, a new virtual adversarial training method is used for fine-tuning to improve models' generalization.  
We show that these techniques significantly improve the efficiency of model pre-training and the performance of both natural language understand (NLU) and natural langauge generation (NLG) downstream tasks. 
Compared to RoBERTa-Large, a DeBERTa model 
% with 45\% model parameters
trained on half of the training data
%computational capacity 
% of the amount used to train the RoBERTa-Large model, our DeBERTa-Large model 
performs consistently better on a wide range of NLP tasks, 
% e.g., 
achieving improvements on MNLI by +0.9\% (90.2\% vs. 91.1\%), on SQuAD v2.0 by +2.3\% (88.4\% vs. 90.7\%) and RACE by +3.6\% (83.2\% vs. 86.8\%). 
Notably, we scale up DeBERTa by training a larger version that consists of 48 Transform layers with 1.5 billion parameters. 
The significant performance boost makes the single DeBERTa model surpass the human performance on the SuperGLUE benchmark~\citep{wang2019superglue} for the first time in terms of macro-average score (89.9 versus 89.8), and the ensemble DeBERTa model sits atop the SuperGLUE leaderboard as of January 6, 2021, outperforming the human baseline by a decent margin (90.3 versus 89.8). The pre-trained DeBERTa models and the source code were released at: \url{https://github.com/microsoft/DeBERTa}\footnote{Our code and models are also available at HuggingFace Transformers: \url{https://github.com/huggingface/transformers}, \url{https://huggingface.co/models?filter=deberta} }.
%We will release the DeBERTa models and the source code to the public.

%The substantially outperforms Google's T5 with 11 billion parameters on the SuperGLUE benchmark \citep{wang2019superglue} and, for the first time, surpasses the human performance (89.9 vs. 89.8). 

% The relative contributions of the two techniques are validated by an ablation study.
%Meanwhile, extensive ablation study is presented to validate the two contributions introduced by DeBERTa.
%Moreover, we continue to run a series of elaborate ablation studies to clearly demonstrate the merits of the two innovations introduced by DeBERTa.  
%The DeBERTa code and pre-trained models will be made publicly available. 

%Recent progress on this area such as BERT \cite{devlin2018bert}, XLNet \cite{yang2019xlnet}, RoBERTa \cite{liu2019roberta} push the pre-training to a new stage with the successfully applying of transformer and masked language model. However, they usally requires a lot of training data, parameters and a lot of training steps which means there are huge space to improve their efficiency. In this work, we explored several ways to improve the efficiency of the per-training of MLM with significant reduced training time and data requirement while the result is much better than previous results.
\end{abstract}

\section{Introduction}
The Transformer has become the most effective neural network architecture for neural language modeling. 
Unlike recurrent neural networks (RNNs) that process text in sequence, Transformers apply self-attention to compute in parallel every word from the input text an attention weight that gauges the influence each word has on another, thus allowing for much more parallelization than RNNs for large-scale model training \citep{vaswani2017attention}. 
Since 2018, we have seen the rise of a set of large-scale Transformer-based Pre-trained Language Models (PLMs), such as 
GPT \citep{radford2019language,brown2020language},
%GPT-2 \citep{radford2019gpt2},
%GPT-3 \citep{brown2020language}, 
BERT \citep{devlin2018bert}, 
RoBERTa \citep{liu2019roberta}, 
XLNet \citep{yang2019xlnet}, 
UniLM \citep{dong2019unilm},
ELECTRA \citep{clark2020electra},
T5 \citep{raffel2019t5},
ALUM \citep{liu2020alum},
StructBERT \citep{wang2019structbert} and
ERINE \citep{sun2019ernie}
. 
These PLMs have been fine-tuned using task-specific labels and created new state of the art in
many downstream natural language processing (NLP) tasks \citep{liu2019mt-dnn,minaee2020deep, jiang2019smart,he2019hnn,he2019x,shen2020exploiting}. 

In this paper, we propose a new Transformer-based neural language model \textbf{DeBERTa} (\textbf{D}ecoding-\textbf{e}nhanced \textbf{BERT} with disentangled \textbf{a}ttention), which 
improves previous state-of-the-art PLMs using two novel techniques: a disentangled attention mechanism, and an enhanced mask decoder.
%, and a virtual adversarial training method for fine-tuning.
%which has proven more effective than RoBERTa and BERT and after fine-tuning leads to better results on a wide range of NLP tasks. 
%DeBERTa makes two modifications to the BERT model.

\paragraph{Disentangled attention.}
% DeBERTa uses a disentangled self-attention mechanism. 
Unlike BERT where each word in the input layer is represented using a vector which is the sum of its word (content) embedding and position embedding, each word in DeBERTa is represented using two vectors that encode its content and position, respectively, and the attention weights among words are computed using disentangled matrices based on their contents and relative positions, respectively. This is motivated by the observation that the attention weight of a word pair depends on not only their contents but their relative positions. For example, the dependency between the words ``deep'' and ``learning'' is much stronger when they occur next to each other than when they occur in different sentences. 

\paragraph{Enhanced mask decoder.}
Like BERT, DeBERTa is pre-trained using masked language modeling (MLM). MLM is a fill-in-the-blank task, where a model is taught to use the words surrounding a mask token to predict what the masked word should be. DeBERTa uses the content and position information of the context words for MLM. The disentangled attention mechanism already considers the contents and relative positions of the context words, but not the absolute positions of these words, which in many cases are crucial for the prediction.
Consider the sentence “a new store opened beside the new mall” with the italicized words “store” and “mall” masked for prediction. Although the local contexts of the two words are similar, they play different syntactic roles in the sentence. (Here, the subject of the sentence is “store” not “mall,” for example.) These syntactical nuances depend, to a large degree, upon the words’ absolute positions in the sentence, and so it is important to account for a word’s absolute position in the language modeling process. DeBERTa incorporates absolute word position embeddings right before the softmax layer where the model decodes the masked words based on the aggregated contextual embeddings of word contents and positions.

In addition, we propose a new virtual adversarial training method for fine-tuning PLMs to downstream NLP tasks. The method is effective in improving models’ generalization.

%\paragraph{Scale invariant fine-tuning.} 
%Virtual adversarial training is a regularization method for improving models’ generalization~\citep{miyato2018vat,jiang2019smart}. 
%It does so by improving a model’s robustness to adversarial examples, which are created by making small perturbations to the input. The model is regularized so that when given a task-specific example, the model produces the same output distribution as it produces on an adversarial perturbation of that example. For NLP tasks, the perturbation is applied to the word embedding instead of the original word sequence. However, the value ranges (norms) of the embedding vectors vary among different words and models. The variance gets larger for bigger models with billions of parameters, leading to some instability of adversarial training. Inspired by layer normalization~\citep{ba2016layer}, to improve the training stability, we developed a Scale-Invariant-Fine-Tuning (SiFT) method where the perturbations are applied to the \emph{normalized} word embeddings.

We show through a comprehensive empirical study that these techniques substantially improve the efficiency of pre-training and the performance of downstream tasks. 
In the NLU tasks, compared to RoBERTa-Large, a DeBERTa model trained on half the training data performs consistently better on a wide range of NLP tasks, 
achieving improvements on MNLI by +0.9\% (90.2\% vs. 91.1\%), on SQuAD v2.0 by +2.3\%(88.4\% vs. 90.7\%), and RACE by +3.6\% (83.2\% vs. 86.8\%). 
In the NLG tasks, DeBERTa reduces the perplexity from 21.6 to 19.5 on the Wikitext-103 dataset. 
We further scale up DeBERTa by pre-training a larger model that consists of 48 Transformer layers with 1.5 billion parameters. 
The single 1.5B-parameter DeBERTa model substantially outperforms T5 with 11 billion parameters on the SuperGLUE benchmark \citep{wang2019superglue} by 0.6\%(89.3\% vs. 89.9\%), and surpasses the human baseline (89.9 vs. 89.8) for the first time.
The ensemble DeBERTa model sits atop the SuperGLUE leaderboard as of January 6, 2021, outperforming the human baseline by a decent margin (90.3 versus 89.8).

\section{Background}
\subsection{Transformer}
A Transformer-based language model is composed of % $L$
stacked Transformer blocks \citep{vaswani2017attention}.
Each block contains a multi-head self-attention layer followed by a fully connected positional feed-forward network. 
The standard self-attention mechanism lacks a natural way to encode word position information.  
Thus, existing approaches add  a positional bias to each input word embedding so that each input word is represented by a vector whose value depends on its content and position.
The positional bias can be implemented using absolute position embedding \citep{vaswani2017attention,radford2019language,devlin2018bert} or relative position embedding \citep{huang2018music,yang2019xlnet}.
%a positional bias to mimic this sequential dependency. 
%Two kinds of remedies have been extensively applied. 
%One is the absolute positional bias in the original Transformer paper \citepp{vaswani2017attention}, GPT \citepp{gpt22019}, BERT \citepp{devlin2018bert}, and RoBERTa \citepp{liu2019roberta}, which perform an element-wise summation to blend the positional embedding with the content embedding. 
%The other is the relative positional bias \citepp{huang2018musictf,yang2019xlnet}, which tried to learn separate positional parameters to capture relative positions. 
It has been shown that relative position representations are more effective for natural language understanding and generation tasks \citep{dai2019transformer,shaw2018self}.
The proposed disentangled attention mechanism differs from all existing approaches in that we represent each input word using two separate vectors that encode a word's content and position, respectively, and attention weights among words are computed using disentangled matrices on their contents and relative positions, respectively. 

\subsection{Masked Language Model}

%Transformer is originally proposed as building blocks of the $Encoder$+$Decoder$ framework for seq-to-seq (s2s) tasks.
% under the  $Encoder$+$Decoder$ structure for text generation,
%Recently,  works in Natural Language Understanding (NLU) often leverage its $Encoder$ to pre-train a language model with a self-supervision objective, i.e. the Masked Language Model (MLM) \citepp{devlin2018bert, liu2019roberta, lan2019albert}. 

Large-scale Transformer-based PLMs 
%have shown promising results in many NLP tasks
%\citep{devlin2018bert, liu2019roberta, lan2019albert} 
are typically pre-trained on large amounts of text to learn contextual word representations using a self-supervision objective, known as Masked Language Model (MLM)~\citep{devlin2018bert}. % which aims to predict words conditioned on their context.
Specifically, given a sequence $\bm{X}=\left\{x_i \right\}$, we corrupt it into $\bm{\tilde{X}}$ by masking 15\% of its tokens at random and then train a language model parameterized by $\theta$ to reconstruct $\bm{X}$ by predicting the masked tokens $\tilde{x}$ conditioned on $\bm{\tilde{X}}$: 
%with a portion of it randomly corrupted and denoted as 
%$\left\{\tilde{x}_i \right\}$, we represent the corrupted sequence as $\bm{\tilde{X}}$. The MLM objective is to reconstruct the corrupted tokens $\{\tilde{x}_i\}$ from $\bm{\tilde{X}}$,
\begin{equation}
\label{mlm}
\begin{split}
       \max_{\theta} \text{log} \; p_{\theta}(\bm{X}|\bm{\tilde{X}}) &= \max_{\theta} \sum_{i\in \mathcal{C}}{\text{log} \;  p_{\theta}(\tilde{x}_i=x_i|\bm{\tilde{X}})} %\\ 
       %&=\sum_{i \in K}{\text{log}{\frac{e^{f_{MLM}(h^{o}_i)\cdot e_i }}{\sum_{j}{e^{f_{MLM}(h^{o}_i) \cdot e_j}}}}} \\
       %f_{MLM}(h_i^o) &= LayerNorm(Linear(h_i^o))
\end{split}
\end{equation}
where $\mathcal{C}$ is the index set of the masked tokens in the sequence. The authors of BERT propose to keep 10\% of the masked tokens unchanged, another 10\% replaced with randomly picked tokens and the rest replaced with the \texttt{[MASK]} token.

\section{The DeBERTa Architecture}
\subsection{Disentangled Attention: A Two-Vector Approach to Content and Position Embedding}
%We will use $\left\{H_i \in R^d \right\}$ to denote the hidden state of an input sequence. 
For a token at position $i$ in a sequence, we represent it using two vectors, $\{\bm{H_i}\}$ and $\{\bm{P_{i|j}}\}$, 
%parts, filler, $\{H_i\}$, and role, $\{P_i\}$, 
which represent its content and relative position with the token at position $j$, respectively.
%where $H_i$ represents the content and $P_i$ represents the position in the sequence. 
The calculation of the cross attention score between tokens $i$ and $j$ can be decomposed into four components as
\begin{equation}\label{decomposition}
\begin{split}
A_{i,j} & = \{\bm{H_i},\bm{P_{i|j}}\} \times \{\bm{H_j}, \bm{P_{j|i}}\}^{\intercal} \\
        & = \bm{H_i}\bm{H_j}^{\intercal} + \bm{H_i}\bm{P_{j|i}}^{\intercal} 
        + \bm{P_{i|j}}\bm{H_j}^{\intercal} + \bm{P_{i|j}}\bm{P_{j|i}}^{\intercal}
\end{split}
\end{equation}
%This implies the attention score between any two tokens in the sequence can be disentangled into four components, i.e.
That is, the attention weight of a word pair can be computed as a sum of four attention scores using disentangled matrices on their contents and positions as  
\textit{content-to-content}, \textit{content-to-position}, \textit{position-to-content}, and \textit{position-to-position}
\footnote{In this sense, our model shares some similarity to Tensor Product Representation \citep{smolensky1990ptr,schlag2019enhancing,chen2019natural} where a word is represented using a tensor product of its filler (content) vector and its role (position) vector.}. 
%which is similar to the TPR (Tensor Product Representation) work \cite{smolensky1990ptr}.

Existing approaches to relative position encoding use a separate embedding matrix to compute the relative position bias in computing attention weights~\citep{shaw2018self, huang2018music}. 
% This is similar to only represent the \textit{content-to-position} term, but ignore the other two terms in \eqref{decomposition}. 
This is equivalent to computing the attention weights using only the  content-to-content and content-to-position terms in \eqref{decomposition}.
We argue that the position-to-content term is also important since the attention weight of a word pair depends not only on their contents but on their relative positions, which can only be fully modeled using both the content-to-position and position-to-content terms.
Since we use \emph{relative} position embedding, the position-to-position term does not provide much additional information and is removed from \eqref{decomposition} in our implementation.

%We argue that this simplification is too partial to fully characterize the relative position dependency and can result in a sheer performance loss. Therefore, we propose to disentangle the position and content thoroughly as shown in \eqref{decomposition}. 
%In other words, the new disentangled attention in \eqref{decomposition} extends existing relative attention bias to incorporate both the \textit{position-to-content} and the \textit{position-to-position} terms. Based on our empirical study, we observe the performance improvements from \textit{position-to-position} is marginal since it ignores the contents, thus we consider the first three terms in \eqref{decomposition}.

%While absolute position bias can capture these four components by adding the position bias directly to the content embedding, relative position bias can't be added directly to the input embedding. Instead, as \cite{shaw2018self}, \cite{huang2018music}, they apply relative position bias to the attention score directly. However, existing methods only considered \textit{content-to-position} bias, while ignored the other two parts. We thought this can't fully capture the relative position dependency and thus we propose disentangled attention which extends existing relative attention bias to \textit{position-to-content} and \textit{position-to-position}. In our experiments, we found the signal from \textit{position-to-position} is very weak as it's unrelated to any contents, so we drop this part in following discussion.

%For simplicity, we use single head self-attention to describe disentangled attention. 
Taking single-head attention as an example, the standard self-attention operation~\citep{vaswani2017attention} can be formulated as:  
\begin{align*}
    \bm{Q} =  \bm{H} \bm{W_q} ,    \bm{K} &=  \bm{H} \bm{W_k}, \bm{V} =  \bm{H} \bm{W_v},    \bm{A} = \frac{\bm{Q}\bm{K^{\intercal}}}{\sqrt{d}} \\
    \bm{H_o}&= \text{softmax}(\bm{A})\bm{V}
\end{align*}  
where $\bm{H} \in R^{N \times d}$ represents the input hidden vectors, $\bm{H_o} \in R^{N \times d}$ the output of self-attention, $\bm{W_q},\bm{W_k},\bm{W_v} \in R^{d \times d}$ the projection matrices, $\bm{A} \in R^{N \times N}$ the attention matrix, $N$ the length of the input sequence, and $d$ the dimension of hidden states.

Denote $k$ as the maximum relative distance,
$\delta(i,j) \in [0,2k)$ as the relative distance from token $i$ to token $j$, which is defined as:
\begin{equation}
\label{dist}
\delta(i,j) = \left\{ \begin{array}{rcl}
     0 & \mbox{for} & i-j \leq  -k\\
     2k-1 & \mbox{for} & i-j \geq k \\
     i-j+k & \mbox{others}.
\end{array}\right. 
\end{equation}

%$\bm{Q_{c,i}} \in R^d$ as the query content vector of the %$i$-th token in the sequence, and
%$\bm{K_{r,\delta(i,j)}} \in R^d$ as the key position vector with relative distance $\delta(i,j)$. 
We can represent the disentangled self-attention with relative position bias as \eqref{dis-att}, 
where $\bm{Q_c}, \bm{K_c}$ and $\bm{V_c}$  are the projected content vectors generated using projection matrices $\bm{W_{q,c}},\bm{W_{k,c}} ,\bm{W_{v,c}}\in R^{d \times d}$ respectively, 
$\bm{P} \in R^{2k \times d}$ represents the relative position embedding vectors shared across all layers (i.e., staying fixed during forward propagation), 
and $\bm{Q_r}$ and $\bm{K_r}$ are projected relative position vectors generated using projection matrices $\bm{W_{q,r}},\bm{W_{k,r}} \in R^{d \times d}$, respectively.   
%Following \cite{shaw2018self,huang2018music}, the content-to-position term can be calculate as $\bm{Q_{c,i}}\bm{K_{r,\delta(i,j)}^{\intercal}}$. 
%We now describe the calculation of the position-to-content term.
%For a given position $i$, position-to-content computes the attentive importance of the key content at $j$ with respect to the query position at $i$, thus the relative distance is $\delta(j,i)$, not $\delta(i,j)$. So, the position-to-content term is calculated as $\bm{Q_{r,\delta(j,i)}}\bm{K_{c,j}^{\intercal}}$, where $\bm{Q_{r,\delta(j,i)}} \in R^d$ is the query vector of relative distance $\delta(j,i)$ and $\bm{K_{c,j}^{\intercal}} \in R^d$ is the key content vector of the $j$-th token in the sequence.

%Putting all together, the disentangled self-attention is computed as \eqref{dis-att}.

%Putting together, we disentangle the attention weight $\bm{A}$ into three components in \eqref{dis-att}. 
\begin{equation} \label{dis-att}
\begin{split}
    \bm{Q_c} = \bm{H} \bm{W_{q,c}}, 
    \bm{K_c} &= \bm{H} \bm{W_{k,c}}, 
    \bm{V_c} = \bm{H} \bm{W_{v,c}}, 
    \bm{Q_r} = \bm{P} \bm{W_{q,r}}, 
    \bm{K_r} = \bm{P} \bm{W_{k,r}}\\
    \tilde{A}_{i,j} &= \underbrace{\bm{Q^{c}_{i}}\bm{{K^{c}_{j}}^{\intercal}}}_{\text{(a) content-to-content}}  
    + \underbrace{\bm{Q^{c}_{i}}\bm{{K^{r}_{\delta(i,j)}}^{\intercal}}}_{\text{(b) content-to-position}}
    + \underbrace{\bm{K^{c}_{j}}\bm{{Q^{r}_{\delta(j,i)}}^{\intercal}}}_{\text{(c) position-to-content}} \\
    \bm{H_o} &= \text{softmax}(\frac{\bm{\tilde{A}}}{\sqrt{3d}})\bm{V_c}
\end{split}
\end{equation}
$\tilde{A}_{i,j}$ is the element of attention matrix $\bm{\tilde{A}}$, representing the attention score from token $i$ to token $j$. 
%$\bm{P} \in R^{2k \times d}$ is the relative distance embedding shared across all layers;
%$\bm{W_{q,c}},\bm{W_{k,c}} ,\bm{W_{v,c}}\in R^{d \times d}$ are the projection matrices corresponding to content; 
%$\bm{W_{q,r}},\bm{W_{k,r}} \in R^{d \times d}$ are the projection matrices corresponding to \bm{P};  
$\bm{Q^{c}_{i}}$ is the $i$-th row of $\bm{Q_c}$. %\textcolor{red}{Qci has been introduced above. we shall remove related information in line 97. same to Kr,delta}; 
$\bm{K^{c}_{j}}$ is the $j$-th row of $\bm{K_c}$. 
% $\bm{P} \in R^{2k \times d}$ is a learnable relative distance embedding shared across all layers;  
% $k$ is the maximum relative distance;
$\bm{K^{r}_{\delta(i,j)}}$ is the $\delta(i,j)$-th row of $\bm{K_r}$ with regarding to relative distance $\delta(i,j)$.
$\bm{Q^{r}_{\delta(j,i)}}$ is the $\delta(j,i)$-th row of $\bm{Q_r}$ with regarding to relative distance $\delta(j,i)$. 
Note that we use $\delta(j,i)$ rather than $\delta(i,j)$ here. This is because for a given position $i$, position-to-content computes the attention weight of the key content at $j$ with respect to the query position at $i$, thus the relative distance is $\delta(j,i)$. 
The position-to-content term is calculated as $\bm{K^{c}_{j}}\bm{{Q^{r}_{\delta(j,i)}}^{\intercal}}$.
%, where $\bm{Q_{r,\delta(j,i)}} \in R^d$ is the query vector of relative distance $\delta(j,i)$ and $\bm{K_{c,j}^{\intercal}} \in R^d$ is the key content vector of the $j$-th token in the sequence. 
The content-to-position term is calculated in a similar way.

Finally, we apply a scaling factor of $\frac{1}{\sqrt{3d}}$ on $\bm{\tilde{A}}$.
The factor is important for stabilizing model training~\citep{vaswani2017attention}, especially for large-scale PLMs. 
%Different from previous relative attention work, to keep the variance invariant, like the scale factor in self-attention, we apply a scale factor as:
%Following the original Transformer model
% Inspired by \cite{vaswani2017attention}, 
%In \eqref{dis-att}, we also apply a scaling factor of $\frac{1}{\sqrt{3d}}$ on $\bm{\tilde{A}}$ which is important for stabilizing model training \cite{vaswani2017attention}, especially for large-scale PLMs.

%, not $\delta(i,j)$. 
%Thus, 

\begin{algorithm}[H]
\caption{Disentangled Attention}
\label{DA}
%\SetAlgoLined
 \begin{algorithmic}[1]
    \INPUT Hidden state $\bm{H}$, relative distance embedding $\bm{P}$, relative distance matrix $\bm{\delta}$.
    Content projection matrix $\bm{W_{k,c}}$, $\bm{W_{q,c}}$, $\bm{W_{v,c}}$,
    position projection matrix $\bm{W_{k,r}}$, $\bm{W_{q,r}}$. 

    \State {$\bm{K_c} = \bm{H}\bm{W_{k,c}}$, $\bm{Q_c} = \bm{H}\bm{W_{q,c}}$, $\bm{V_c} = \bm{H}\bm{W_{v,c}}$,   $\bm{K_r} = \bm{P}\bm{W_{k,r}}$, $\bm{Q_r} = \bm{P}\bm{W_{q,r}}$}
    \State {$\bm{A_{c\rightarrow c}} = \bm{Q_c K_c^{\intercal}}$} 
    \For{$i=0,...,N-1$} 
        \State {$\bm{\tilde{A}_{c\rightarrow p}}[i,:] = \bm{Q_{c}}[i,:] \bm{K_r^{\intercal}}$} 
    \EndFor
    \For {$i=0,...,N-1$}
        \For {$j=0,...,N-1$}
        \State {$\bm{A_{c\rightarrow p}}[i,j] = \bm{\tilde{A}_{c\rightarrow p}}[i,\bm{\delta}[i,j]]$} 
        \EndFor
    \EndFor
    \For{$j=0,...,N-1$}
        \State {$\bm{\tilde{A}_{p\rightarrow c}}[:,j] = \bm{K_{c}}[j,:] \bm{Q_r^{\intercal}}$} 
    \EndFor
    \For {$j=0,...,N-1$}
        \For {$i=0,...,N-1$}
        \State {$\bm{A_{p\rightarrow c}}[i,j] = \bm{\tilde{A}_{p\rightarrow c}}[\bm{\delta}[j,i],j]$} 
        \EndFor
    \EndFor
    \State {$\bm{\tilde{A}}=\bm{A_{c\rightarrow c}} + \bm{A_{c\rightarrow p}} + \bm{A_{p\rightarrow c}}$}
    \State {$\bm{H_o} = \text{softmax}(\frac{\bm{\tilde{A}}}{\sqrt{3d}})\bm{V_c}$}
    \OUTPUT $\bm{H_o}$
 \end{algorithmic}

\end{algorithm}
\subsubsection{Efficient implementation}
For an input sequence of length $N$, it requires a space complexity of $O(N^2d)$ \citep{shaw2018self,huang2018music,dai2019transformer} to store the relative position embedding for each token. 
%Since relative distance vectors vary from query to query, if we directly multiply a query vector by its corresponding relative position embedding, this will unavoidably let the $d$-dimension vector multiply a $N\times d$ matrix for $N$ times.
%\textcolor{red}{could we map to an equation we are talking about here? the last term in Equation 3? i also don't understand what you mean of multiplied with a N * d matrix}. 
%The memory cost of this multiplication will be $O(N^2d)$ since we need to allocate the memory to store the relative distance embedding which has a size of $N \times N \times d$  . 
However, taking content-to-position as an example, we note that since $\delta(i,j) \in [0,2k)$ and the embeddings of all possible relative positions are always a subset of $\bm{K_r} \in R^{2k \times d}$, then we can reuse $\bm{K_r}$ in the attention calculation for all the queries.

In our experiments, we set the maximum relative distance $k$ to 512 for pre-training.
The disentangled attention weights can be computed efficiently using Algorithm~\ref{DA}.
Let $\bm{\delta}$ be the relative position matrix according to \eqref{dist}, i.e., $\bm{\delta}[i,j]=\delta(i,j)$.
Instead of allocating a different relative position embedding matrix for each query, we multiply each $query$ vector $\bm{Q_c}[i,:]$ by $\bm{K_r}^\intercal \in R^{d \times 2k}$, as in line $3-5$. 
Then, we extract the attention weight using the relative position matrix $\bm{\delta}$ as the index, as in line $6-10$. 
To compute the position-to-content attention score, we  calculate $\bm{\tilde{A}_{p\rightarrow c}}[:,j]$, i.e., the column vector of the attention matrix $\bm{\tilde{A}_{p\rightarrow c}}$, by multiplying each $key$ vector $\bm{K_c}[j,:]$ by $\bm{Q_r}^\intercal$, as in line $11-13$. 
Finally, we extract the corresponding attention score via the relative position matrix $\bm{\delta}$ as the index, as in line $14-18$. 
In this way, we do not need to allocate memory to store a relative position embedding for each query and thus reduce the space complexity to $O(kd)$ (for storing $\bm{K_r}$ and $\bm{Q_r}$). 
\subsection{Enhanced Mask Decoder Accounts for Absolute Word Positions}
\label{EMD}
%The DeBERTa model has two additional extensions. One is to address a limitation of the relative positions which have been fully captured by the disentangled attentions. The other is to enable generation tasks and a multi-task learning objective.

DeBERTa is pretrained using MLM, where a model is trained to use the words surrounding a mask token to predict what the masked word should be. 
DeBERTa uses the content and position information of the context words for MLM. The disentangled attention mechanism already considers the contents and relative positions of the context words, but not the absolute positions of these words, which in many cases are crucial for the prediction.

Given a sentence ``a new \textbf{store} opened beside the new \textbf{mall}'' with the words ``store'' and ``mall'' masked for prediction. Using only the local context (e.g., relative positions and surrounding words) is insufficient for the model to distinguish \textit{store} and \textit{mall} in this sentence, since both follow the word \textit{new} with the same relative positions. 
To address this limitation, the model needs to take into account absolute positions, as complement information to the relative positions. 
For example, the subject of the sentence is “store” not “mall”. These syntactical nuances depend, to a large degree, upon the words’ absolute positions in the sentence.

There are two methods of incorporating absolute positions. The BERT model incorporates absolute positions in the input layer. 
In DeBERTa, we %propose an alternative to consider it
incorporate them right after all the Transformer layers but before the \emph{softmax} layer for masked token prediction, as shown in Figure \ref{fig:emd}. 
In this way, DeBERTa captures the relative positions in all the Transformer layers and only uses absolute positions as complementary information when decoding the masked words. % \textit{softmax} decoding layer. 
Thus, we call DeBERTa's decoding component an Enhanced Mask Decoder (EMD). 
In the empirical study, we compare these two methods of incorporating absolute positions and observe that EMD works much better. 
We conjecture that the early incorporation of absolute positions used by BERT might undesirably hamper the model from learning sufficient information of relative positions. In addition, EMD also enables us to introduce other useful information, in addition to positions, for pre-training. We leave it to future work.

\section{Scale Invariant Fine-Tuning}

This section presents a new virtual adversarial training algorithm, Scale-invariant-Fine-Tuning (SiFT), a variant to the algorithm described in~\cite{miyato2018vat,jiang2019smart}, for fine-tuning. 

Virtual adversarial training is a regularization method for improving models’ generalization. It does so by improving a model’s robustness to adversarial examples, which are created by making small perturbations to the input. The model is regularized so that when given a task-specific example, the model produces the same output distribution as it produces on an adversarial perturbation of that example. 

For NLP tasks, the perturbation is applied to the word embedding instead of the original word sequence. However, the value ranges (norms) of the embedding vectors vary among different words and models. The variance gets larger for bigger models with billions of parameters, leading to some instability of adversarial training. 

Inspired by layer normalization~\citep{ba2016layer}, we propose the SiFT algorithm that improves the training stability by applying the perturbations to the \emph{normalized} word embeddings. 
Specifically, when fine-tuning DeBERTa to a downstream NLP task in our experiments, SiFT first normalizes the word embedding vectors into stochastic vectors, and then applies the perturbation to the normalized embedding vectors. We find that the normalization substantially improves the performance of the fine-tuned models. The improvement is more prominent for larger DeBERTa models. 
Note that we \textbf{only} apply SiFT to  {\ModelName}$_{1.5B}$ on SuperGLUE tasks in our experiments and we will provide a more comprehensive study of SiFT in our future work.

\section{Experiment}
%In this section, we present a comprehensive study of {\ModelName} compared with a list on pre-trained models including BERT \cite{devlin2018bert}, RoBERTa \cite{liu2019roberta} and XLNet \cite{yang2019xlnet}. Then, we show the efficiency and effectiveness of  {\ModelName} on a wide spectrum of Natural Language Understanding benchmarks including GLUE \cite{wang2018glue}, SuperGLUE \cite{superglue2019_8589}, SQuAD v1.1 \cite{squad1}, SQuAD v2.0 \cite{squad2}, RACE, SWAG \cite{zellers2018swag} and Named-entity recognition (NER) in OntoNotes 5.0 \cite{ontonotes5}. For details of the benchmarks, please see Appendix ~\ref{sec:appendix}. 
This section reports {\ModelName} results on various NLU tasks. 
\subsection{Main Results on NLU tasks}
\label{subsec:main}
Following previous studies of PLMs,
% papers on BERT, RoBERTa and XLNet, 
we report results using large and base models.
%size with a comparison to all of them. 
\subsubsection{Performance on Large Models}
\begin{table*}[htb!]
    \centering
    \begin{tabular}{@{\hskip3pt}l@{\hskip2pt}|@{\hskip2pt} c@{\hskip2pt}| @{\hskip2pt}c@{\hskip2pt}|c@{\hskip2pt}|c@{\hskip2pt}|c@{\hskip2pt}|c@{\hskip2pt}|c@{\hskip2pt}|c@{\hskip2pt}|c@{\hskip2pt}}
        \toprule
        \multirow{2}{*}{\bf Model} & {CoLA} &{QQP} &{MNLI-m/mm} &SST-2 &STS-B&QNLI&RTE&MRPC& Avg.\\ 
        & Mcc & Acc & Acc & Acc &Corr&Acc&Acc&Acc\\
        \midrule
        BERT$_{large}$ &60.6 &91.3  & 86.6/- & 93.2 &90.0&92.3&70.4&88.0 &84.05 \\ \hline
        RoBERTa$_{large}$ &68.0 & 92.2  & 90.2/90.2 & 96.4 &92.4&93.9&86.6&90.9& 88.82 \\ \hline
        XLNet$_{large}$& 69.0 & 92.3  & 90.8/90.8 & \textbf{97.0} &92.5&94.9&85.9&90.8 & 89.15 \\ \hline
%ALBERT$_{xxlarge}$&  \textbf{71.4} & 92.2  & 90.8/- & 96.9 &\textbf{93.0}&95.3&\textbf{89.2}&90.9 &89.96 \\ \hline

ELECTRA\textsubscript{large}&69.1 & \textbf{92.4} &90.9/- &96.9& 92.6&95.0 & 88.0&90.8 &89.46 \\ \hline
        {\ModelName}$_{large}$ &\textbf{70.5} & 92.3 & \textbf{91.1/91.1} &96.8 & \textbf{92.8} &\textbf{95.3}&\textbf{88.3}& \textbf{91.9} &\textbf{90.00}\\
        \bottomrule
        \end{tabular}
    \caption{
    Comparison results on the GLUE development set. 
    }
    \label{tab:glue}
    \vspace{-2mm}
\end{table*}

% avg scores
% 60.6	91.3	86.6	93.2	90	92.3	70.4	88	84.05
% 68	92.2	90.2	96.4	92.4	93.9	86.6	90.9	88.825
% 69	92.3	90.8	97	92.5	94.9	85.9	90.8	89.15
% 71.4	92.2	90.8	96.9	93	95.3	89.2	90.9	89.9625
% 69.1	92.4	90.9	96.9	92.6	95	88	90.8	89.4625
% 70.5	92.3	91.2	96.7	92.5	95.3	88.1	93.4	90

% \begin{table*}[htb!]
%     \centering
%     \begin{tabular}{@{\hskip3pt}l@{\hskip2pt}|@{\hskip2pt} c@{\hskip2pt}| @{\hskip2pt}c@{\hskip2pt}|c@{\hskip2pt}|c@{\hskip2pt}|c@{\hskip2pt}|c@{\hskip2pt}|c@{\hskip2pt}|c@{\hskip2pt}|@{\hskip2pt}c@{\hskip2pt}}
%         \toprule
%         \multirow{2}{*}{\bf Model} &{\bf Model} & {CoLA} &{QQP} &{MNLI-m/mm} &SST-2 &STS-B&QNLI&RTE&MRPC\\ 
%         &{\bf Size}& Mcc & F1/Acc & Acc & Acc &Corr&Acc&Acc&Acc\\
%         \midrule
%         BERT$_{large}$ &335M &60.6 &91.3  & 86.6/- & 93.2 &90.0&92.3&70.4&88.0 \\ \hline
%         RoBERTa$_{large}$& 355M& 68.0 & 92.2  & 90.2/90.2 & 96.4 &92.4&93.9&86.6&90.9 \\ \hline
%         XLNet$_{large}$& 340M& 69.0 & 92.3  & 90.8/90.8 & \textbf{97.0} &92.5&94.9&85.9&90.8 \\ \hline
% ALBERT$_{large}$& 340M& 69.0 & 92.3  & 90.8/90.8 & \textbf{97.0} &92.5&94.9&85.9&90.8 \\ \hline        
%         {\ModelName}$_{large}$ &390M &\textbf{69.5} & \textbf{92.3} & \textbf{91.1/91.1} & 96.5 & \textbf{92.5} &\textbf{95.3} &\textbf{88.1}& \textbf{92.5}\\
%         \bottomrule
%         \end{tabular}
%     \caption{
%     Comparison results on the GLUE development set. 
%     }
%     \label{tab:glue}
% \end{table*}

We pre-train our large models following the setting of BERT~\citep{devlin2018bert}, except that we use the BPE vocabulary of ~\cite{radford2019language,liu2019roberta}.
For training data, we use Wikipedia (English Wikipedia dump\footnote{https://dumps.wikimedia.org/enwiki/}; 12GB), BookCorpus~\citep{bookcorpus} (6GB), OPENWEBTEXT (public Reddit content~\citep{Gokaslan2019OpenWeb}; 38GB), and STORIES (a subset of CommonCrawl~\citep{trinh2018simple}; 31GB). 
The total data size after data deduplication~\citep{shoeybi2019megatron} is about 78G. 
Refer to Appendix A.2 for a detailed description of the pre-training dataset.

We use 6 DGX-2 machines (96 V100 GPUs) to train the models. 
A single model trained with 2K batch size and 1M steps takes about 20 days.
Refer to Appendix \ref{sec:appendix} for the detailed hyperparamters. 
% We use a batch size of 2048. Our training set contains Wikipedia \footnote{https://dumps.wikimedia.org/enwiki}, bookcorpus \cite{bookcorpus}, openwebtext \cite{Gokaslan2019OpenWeb} and STORIES (a subset of CommonCrawl) \cite{trinh2018simple}. The total data size after data deduplication is about 76GB. We set the learning rate as 2e-4, with the number of warm-up steps as 10000. Following RoBERTa \cite{liu2019roberta}, we  adopted dynamic data batching. We also include span masking\cite{joshi2019spanbert} as the additional masking strategy with the span size up to 3. For a single complete training, we use 6 DGX-2 machines with 96 V100 GPUs to train the model for 1M steps and it takes about 20 days. For fine-tuning, we choose a learning rate among $\{5e-6, 8e-6, 9e-6, 1e-5\}$ and train each of them up to 8 epochs.
% We pick the best model according to the performance on the task-specific dev sets.

We summarize the results on eight NLU tasks of GLUE~\citep{wang2018glue} in Table~\ref{tab:glue}, where DeBERTa is compared {\ModelName} with previous Transform-based PLMs of similar structures (i.e. 24 layers with hidden size of 1024) including BERT, RoBERTa, XLNet, ALBERT and ELECTRA.
Note that RoBERTa, XLNet and ELECTRA are pre-trained on 160G training data while {\ModelName} is pre-trained on 78G training data. RoBERTa and XLNet are pre-trained for 500K steps with 8K samples in a step, which amounts to four billion training samples. 
{\ModelName} is pre-trained for one million steps with 2K samples in each step.  
This amounts to two billion training samples, approximately half of either RoBERTa or XLNet.  
Table~\ref{tab:glue} shows that compared to BERT and RoBERTa, {\ModelName} performs consistently better across all the tasks. 
Meanwhile, {\ModelName} outperforms XLNet in six out of eight tasks. 
Particularly, the improvements on MRPC (1.1\% over XLNet and 1.0\% over RoBERTa), RTE (2.4\% over XLNet and 1.7\% over RoBERTa) and CoLA (1.5\% over XLNet and 2.5\% over RoBERTa) are significant. 
{\ModelName} also outperforms other SOTA PLMs, i.e., ELECTRA\textsubscript{large} and XLNet\textsubscript{large}, in terms of average GLUE score.  
%, ALBERT\textsubscript{xxlarge} \footnote{ The hidden dimension of ALBERT\textsubscript{xxlarge} is 4 times of DeBERTa and the computation cost is about 4 times of DeBERTa. } and 

Among all GLUE tasks, MNLI is most often used as an indicative task to monitor the research progress of PLMs. 
{\ModelName} significantly outperforms all existing PLMs of similar size on MNLI and creates a new state of the art. % for large models. 
\begin{table*}[htb!]
    \centering
    \begin{tabular}{@{\hskip2pt}l@{\hskip3pt}|@{\hskip2pt} c@{\hskip2pt}|| @{\hskip2pt}c@{\hskip2pt}@{\hskip2pt}c@{\hskip2pt}|@{\hskip2pt}c@{\hskip2pt}|@{\hskip2pt}c@{\hskip2pt}||@{\hskip2pt}c@{\hskip2pt}||@{\hskip2pt}c@{\hskip2pt}}
        \toprule
        \multirow{2}{*}{\bf Model} &{MNLI-m/mm} & {SQuAD v1.1} &{SQuAD v2.0} &RACE &ReCoRD  &SWAG & NER\\ 
        & Acc & F1/EM & F1/EM &Acc&F1/EM& {Acc}   & F1\\
        \midrule
        BERT$_{large}$ & 86.6/-& 90.9/84.1 &81.8/79.0  &72.0 &-   &86.6 &92.8\\ \hline
        ALBERT$_{large}$ & 86.5/-& 91.8/85.2 & 84.9/81.8 &75.2 &-  &-&- \\ \hline
        RoBERTa$_{large}$ & 90.2/90.2& 94.6/88.9 & 89.4/86.5 &83.2 &90.6/90.0   &89.9 &93.4\\ \hline
        XLNet$_{large}$ & 90.8/90.8& 95.1/89.7 & 90.6/87.9 &85.4 &-  &-&- \\ \hline
        Megatron\textsubscript{336M} & 89.7/90.0 & 94.2/88.0 & 88.1/84.8 &83.0 &-   &-&- \\ \hline
        {\ModelName}$_{large}$ & \textbf{91.1/91.1} &\textbf{95.5/90.1} & \textbf{90.7/88.0} & \textbf{86.8} &\textbf{91.4/91.0}   &\textbf{90.8} &\textbf{93.8}\\ \hline \hline
        ALBERT$_{xxlarge}$ & 90.8/-& 94.8/89.3 & 90.2/87.4 &86.5 &-  &-&- \\ \hline
        Megatron\textsubscript{1.3B} & 90.9/91.0 & 94.9/89.1 & 90.2/87.1 & 87.3 &- &-&- \\ \hline
        Megatron\textsubscript{3.9B} & 91.4/91.4 & 95.5/90.0 & 91.2/88.5 &89.5&-  &-&- \\      
        \bottomrule
        \end{tabular}
    \caption{
    Results on MNLI in/out-domain,  SQuAD v1.1, SQuAD v2.0, RACE, ReCoRD, SWAG, CoNLL 2003 NER development set. Note that missing results in literature are signified by ``-''.}
    \label{tab:large}
    \vspace{-4mm}
\end{table*}

In addition to GLUE, DeBERTa is evaluated on three categories of NLU benchmarks: 
(1) Question Answering: SQuAD v1.1 \citep{squad1}, SQuAD v2.0 \citep{squad2}, RACE \citep{lai2017race}, ReCoRD \citep{zhang2018record} and SWAG \citep{zellers2018swag}; 
(2) Natural Language Inference: MNLI \citep{mnli2018}; and 
(3) NER: CoNLL-2003.
For comparison, we include 
ALBERT\textsubscript{xxlarge}~\citep{lan2019albert}
\footnote{ The hidden dimension of ALBERT\textsubscript{xxlarge} is 4 times of DeBERTa and the computation cost is about 4 times of DeBERTa.} and 
Megatron~\citep{shoeybi2019megatron} with three different model sizes, denoted as Megatron\textsubscript{336M}, Megatron\textsubscript{1.3B} and Megatron\textsubscript{3.9B}, respectively, which are trained using the same dataset as RoBERTa. 
Note that Megatron\textsubscript{336M} has a similar model size as other models mentioned above\footnote{T5~\citep{raffel2019t5} has more parameters (11B). \cite{raffel2019t5} only report the test results of T5 which are not comparable with other models.}. 

We summarize the results in Table~{\ref{tab:large}}. 
Compared to the previous SOTA PLMs with a similar model size (i.e., BERT, RoBERTa, XLNet,  ALBERT\textsubscript{large}, and Megatron\textsubscript{336M}),  {\ModelName} shows superior performance in all seven tasks. 
Taking the RACE benchmark as an example, {\ModelName} significantly outperforms XLNet by +1.4\% (86.8\% vs. 85.4\%). 
Although Megatron\textsubscript{1.3B} is three times larger than {\ModelName}, {\ModelName} outperforms it in three of the four benchmarks. We further report {\ModelName} on text generation tasks in Appendix~\ref{subsec:generation}.
%All the results show the superior performance of {\ModelName} in various downstream tasks. 
% We are confident that 
% {\ModelName} could have performed even better with a larger model size. We leave it to future work.
%and share larger models in its github repository once available.

\subsubsection{Performance on Base Models}
Our setting for base model pre-training is similar to that for large models. 
The base model structure follows that of the BERT base model, i.e., $L=12, H=768, A=12$. 
We use 4 DGX-2 with 64 V100 GPUs to train the base model. 
It takes 10 days to finish a single pre-training of 1M training steps with batch size 2048. 
We train {\ModelName} using the same 78G dataset, and compare it to RoBERTa and XLNet trained on 160G text data. 
%For a detailed comparison of datasets for pre-training, please refer to Appendix A.2.  
%Refer to Appendix~\ref{subsec:details} for detailed hyperparameters used for pre-training and fine-tuning.
%The other difference with the Large model is that we use a different set of learning rates in the task-specific model fine-tuning, which are $\{2e-5, 3e-5,4e-5\}$. 

We summarize the base model results in Table~\ref{tab:base-dev}. 
Across all three tasks, {\ModelName} consistently outperforms RoBERTa and XLNet by a larger margin than that in large models.  
For example, on MNLI-m, {\ModelName\textsubscript{base}} obtains +1.2\%  (88.8\% vs. 87.6\%) over RoBERTa\textsubscript{base}, and +2\% (88.8\% vs. 86.8\%) over XLNet\textsubscript{base}.
%Simultaneously, 
% \ModelName\textsubscript{base} achieves the best performance on all the tasks among all competing base models.
% of around 120M parameters. 
% This significant and consistent improvements continue to echo the great performance of {\ModelName}.   
\begin{table*}[htb!]
    \centering
    \begin{tabular}{@{\hskip2pt}l| c | c | c @{\hskip2pt}}
        \toprule
%        \multirow{2}{*}{\bf Model} & {SQuAD v1.1} &{SQuAD v2.0} &{MNLI-m/mm} \\ 
        {\bf Model} &{MNLI-m/mm (Acc)} & {SQuAD v1.1 (F1/EM)} &{SQuAD v2.0 (F1/EM)}  \\ 
       % & F1/EM & F1/EM & { Acc} \\
        \midrule
%        BERT$_{base}$ \cite{devlin2018bert} & 84.3/84.7 & 88.5/80.8 &76.3/73.7    \\ \hline
        RoBERTa$_{base}$ & 87.6/- & 91.5/84.6 & 83.7/80.5  \\ \hline
        XLNet$_{base}$ & 86.8/- & -/- & -/80.2  \\ \hline
        {\ModelName}$_{base}$ & \textbf{88.8/88.5}  &\textbf{93.1/87.2} & \textbf{86.2/83.1}  \\
        \bottomrule
        \end{tabular}
    \caption{
    Results on MNLI in/out-domain (m/mm), SQuAD v1.1 and v2.0   development set. 
    }
    \label{tab:base-dev}
    \vspace{-3mm}
\end{table*}

\subsection{Model Analysis}
%In this section, we focus on analyzing the factors leading to the good performance of {\ModelName}.  
In this section, we first present an ablation study to quantify the relative contributions of different components introduced in {\ModelName}. 
Then, we study the convergence property to characterize the model training efficiency. 
%Due to the high training cost and long training time, it is infeasible for us to run many analysis experiments with the same setting as previous section. Therefore, 
We run experiments for analysis using the base model setting: a model is pre-trained using 
the Wikipedia + Bookcorpus dataset 
for 1M steps with batch size 256 
in 7 days on a DGX-2 machine with 16 V-100 GPUs.
Due to space limit, we visualize the different attention patterns of {\ModelName} and RoBERTa in Appendix A.7. 
\subsubsection{Ablation study}
\label{subsec:abs}
%Here, we study the importance of each proposed component of {\ModelName} on four NLU tasks. Due to the limitation of time and computational resources, all the analysis is based on the base model which has 125M parameters. For a fair comparison, all the models are trained on Wikipedia and BookCorpus:
%Before we perform any ablation study, we want to ensure that there are minimal side effects impacting the credibility of the experimental results. For this purpose, 
To verify our experimental setting, we pre-train the RoBERTa base model from scratch.  
%following {\ModelName}\textsubscript{base}. 
The re-pre-trained RoBERTa model is denoted as RoBERTa-ReImp\textsubscript{base}. 
%and treat it as a fair baseline for {\ModelName}. 
To investigate the relative contributions of different components in {\ModelName}, we develop three variations:
%of the original {\ModelName} model, i.e.:
\begin{itemize}
%    \item RoBERTa\textsubscript{base} is a RoBERTa base model from \cite{liu2019roberta}.
%    \item RoBERTa-ReImp\textsubscript{base} is our re-implemented model with additional span masking \cite{joshi2019spanbert}. Note that all our models are built upon this implementation.
%    \item {\ModelName\textsubscript{base}} is the proposed {\ModelName} base model.
    \item {-\DecoderName} is the {\ModelName} base model without \DecoderName.
    \item {-C2P} is the  {\ModelName} base model without the content-to-position term ((c) in Eq.~\ref{dis-att}).
    \item {-P2C} is the {\ModelName} base model without the position-to-content term ((b) in Eq.~\ref{dis-att}). As XLNet also uses the relative position bias, this model is close to XLNet plus \DecoderName.
\end{itemize}
\begin{table*}[htb!]
    \centering
    \begin{tabular}{@{\hskip1pt}l| c| cc|c@{\hskip1pt}}
        \toprule
        \multirow{2}{*}{\bf Model} &{MNLI-m/mm}& {SQuAD v1.1} &{SQuAD v2.0} &RACE\\ 
         & { Acc}& F1/EM & F1/EM  & Acc\\
        \midrule
        BERT$_{base}$ \cite{devlin2018bert} & 84.3/84.7 & 88.5/81.0 &76.3/73.7   & 65.0 \\ %\hline        
        RoBERTa$_{base}$ \cite{liu2019roberta} & 84.7/- & 90.6/- &79.7/-   & 65.6\\ %\hline
        XLNet$_{base}$ \cite{yang2019xlnet}& 85.8/85.4 & -/- & 81.3/78.5  & 66.7\\ \hline
        RoBERTa-ReImp$_{base}$ & 84.9/85.1 & 91.1/84.8 &79.5/76.0   & 66.8\\ \hline %base+MK+SPAN
        {\ModelName}$_{base}$ & \textbf{86.3/86.2} &\textbf{92.1/86.1} & \textbf{82.5/79.3}  &\textbf{ 71.7}\\
        -EMD & 86.1/86.1& 91.8/85.8 &81.3/78.0   & 70.3\\ %C2P+P2C+SPAN+MK
        -C2P  & 85.9/85.7 & 91.6/85.8 &81.3/78.3   & 69.3\\ % P2C+SPAN+MK+MSD
        -P2C  & 86.0/85.8 & 91.7/85.7 &80.8/77.6   & 69.6\\ % C2P+SPAN+MK+MSD
        -(EMD+C2P) & 85.8/85.9 & 91.5/85.3 &80.3/77.2   & 68.1\\ % SPAN+MK        
        -(EMD+P2C)  & 85.8/85.8& 91.3/85.1 &80.2/77.1    & 68.5\\ % C2P+SPAN+MK
        \bottomrule
        \end{tabular}
    \caption{
    Ablation study of the DeBERTa base model.
    }
    \label{tab:ablation}
    %\vspace{-3mm}
\end{table*}
%Note that all the models are pre-trained on Wikipedia and BookCorpus for a fair comparison and other training setting is the same as BERT base.
Table~\ref{tab:ablation} summarizes the results on four benchmark datasets. 
%The re-implemented RoBERTa-ReImp\textsubscript{base} obtains the similar performance with the RoBERTa\textsubscript{base} in literature across all the four NLU tasks indicating a fairness of comparison. 
First, RoBERTa-ReImp performs similarly to RoBERTa across all  benchmark datasets, verfiying that our setting is reasonable. 
%Thus, we can confidently treat RoBERTa-ReImp as a solid baseline for comparison.
%for {\ModelName}\textsubscript{base}. 
% This also makes following comparison more truthful. 
Second, we see that removing any one component in {\ModelName} results in a sheer performance drop. 
For instance, removing EMD (-EMD) results in a loss of 1.4\% (71.7\% vs. 70.3\%) on RACE, 0.3\% (92.1\% vs. 91.8\%) on SQuAD v1.1, 1.2\% (82.5\% vs. 81.3\%) on SQuAD v2.0, 0.2\% (86.3\% vs. 86.1\%) and 0.1\% (86.2\% vs. 86.1\%) on MNLI-m/mm, respectively. 
Similarly, removing either \textit{content-to-position} or  \textit{position-to-content} leads to inferior performance in all the benchmarks.
As expected, removing two components results in even more substantial loss in performance. 

\subsection{Scale up to 1.5 billion parameters}
Larger pre-trained models have shown better generalization results \citep{raffel2019t5,brown2020language, shoeybi2019megatron}. 
Thus, we have built a larger version of DeBERTa with 1.5 billion parameters, denoted as DeBERTa$_{1.5B}$.
The model consists of 48 layers with a hidden size of 1,536 and 24 attention heads
\footnote{See Table~\ref{tbl:hyper-pre-train} in Appendix for the model hyperparameters.}.
DeBERTa$_{1.5B}$ is trained on a pre-training dataset amounting to 160G, similar to that in~\cite{liu2019roberta}, with a new vocabulary of size 128K constructed using the dataset.
% More details of the hyperparameters are provided in Table~ \ref{tbl:hyper-pre-train}.
%, and was trained using 160G data with a rebuilt vocabulary size of 128k \citep{liu2019roberta}. The detailed configuration of each model is specified in Table~\ref{tbl:hyper-pre-train}. In order to save the GPU memory, the projection matrices of relative position embedding $W_{k,r},W_{q,r}$ with $W_{k,c},W_{q,c}$ are shared in all attention layers. At last, a convolution layer is added aside with the first transformer layer to induce n-gram knowledge of sub-word encodings and the outputs are summed up to feed to next transformer layer. 
 
To train DeBERTa$_{1.5B}$, we optimize the model architecture as follows. 
First, we share the projection matrices of relative position embedding $W_{k,r},W_{q,r}$ with $W_{k,c},W_{q,c}$, respectively, in all attention layers to reduce the number of model parameters. Our ablation study in Table \ref{tab:v2} on base models shows that the projection matrix sharing reduces the model size while retaining the model performance.
%Second, 
%instead of using 78G data in our large and base model, the DeBERTa\textsubscript{1.5B} model is trained with. 
%we increase the size of the pre-training dataset to 160G, similar to that in \citep{liu2019roberta}, with a new vocabulary of size 128K constructed using the dataset. 
Second, a convolution layer is added aside the first Transformer layer to induce n-gram knowledge of sub-word encodings and their outputs are summed up before feeding to the next Transformer layer~\footnote{Please refer to Table~\ref{tab:xxlarge} in Appendix \ref{subsec:scale} for the ablation study of different model sizes, and Table \ref{tab:v2} in Appendix \ref{subsec:scale} for the ablation study of new modifications.}.

%We also use a new virtual adversarial training algorithm, Scale-Invariant-Fine-Tuning (SiFT), a variant to the algorithm described in~\cite{miyato2018vat,jiang2019smart}, for fine-tuning.
%SiFT is intended to improve model’s generalization using adversarial examples created by making small perturbations to the input. 
%When fine-tuning DeBERTa to a downstream NLU task, the model is regularized so that given a task-specific example the model produces the same output distribution as it produces on an adversarial perturbation of that example. 
%SiFT is derived from \citep{miyato2018vat,jiang2019smart}.
%Unlike existing adversarial training methods that apply perturbations directly on the embeddings of input words, 
%SiFT adds a normalization layer right before the perturbation layer such that
%SiFT first normalizes the word embedding vectors into stochastic vectors, and then applies the perturbation to the normalized embedding vectors. 
%We find that the normalization substantially improves the performance of the fine-tuned models in our experiments. 
%We leave a comprehensive study of SiFT to future work.
% substantially gain over baseline fine-tuning, it needs further study on more datasets to come to a  clear conclusion. We will leave it in our future exploration.

Table~\ref{tab:superglue} reports the test results of SuperGLUE \citep{wang2019superglue} which is one of the most popular NLU benchmarks. 
% of general language understanding 
SuperGLUE consists of a wide of NLU tasks, including  
Question Answering \citep{clark2019boolq, khashabi2018multirc, zhang2018record}, 
Natural Language Inference \citep{rte1,rte2,rte3, rte5}, Word Sense Disambiguation \citep{pilehvar2019wic}, and Reasoning \citep{levesque2011winograd,roemmele2011choice}. 
Since its release in 2019, top research teams around the world have been developing large-scale PLMs that have driven striking performance improvement on SuperGLUE. 

The significant performance boost due to scaling DeBERTa to a larger model makes the single DeBERTa$_{1.5B}$ surpass the human performance on SuperGLUE for the first time in terms of macro-average score (89.9 versus 89.8) as of December 29, 2020, and the ensemble DeBERTa model (DeBERTa$_{Ensemble}$) sits atop the SuperGLUE benchmark rankings as of January 6, 2021, outperforming the human baseline by a decent margin (90.3 versus 89.8). 
Compared to T5, which consists of 11 billion parameters, the 1.5-billion-parameter DeBERTa is much more energy efficient to train and maintain, and it is easier to compress and deploy to apps of various settings.

%Compared to the pre-trained T5 11 billion parameters, DeBERTa\textsubscript{1.5B} outperforms the T5 model, which is the SOTA model containing 11 billion parameters, by 0.6 SuperGLUE score (89.9 vs. 89.3). 
%More importantly, our 
% DeBERTa\textsubscript{1.5B} also surpasses the human performance in overall score (89.9 vs. 89.8) for the first time as of December 29, 2020. 

%Since larger models usually achieve a better performance \citep{raffel2019t5,brown2020language, shoeybi2019megatron}, we scale up the DeBERTa model to XXLarge size with 1.5 billion parameters. Besides the increase of model size, we introduced several improvements to the model structure. 
%to further improve the model efficiency. 
%First, we share projection matrix of relative position embedding $W_{k,r},W_{q,r}$ with $W_{k,c},W_{q,c}$ in all attention layers to save model parameters. Our ablation study on the base-size 12-layer model shows by sharing the projection matrix, the model performance almost keep the same.
%Second, 
%instead of using 78G data in our large and base model, the XXLarge model is trained with 
%we increase the pre-training data to 160G data, similar size as \citep{liu2019roberta}, with a new vocabulary of 128K  built from this new data. Third, a convolution layer is added aside with the first transformer layer to induce n-gram knowledge of sub-word encodings and the outputs are summed up to feed to next transformer layer. The model is trained with the same hyper parameters as our large model in Table \ref{tbl:hyper-pre-train}. 
\begin{table*}[htb!]
    \centering
    \begin{tabular}{@{\hskip2pt}l@{\hskip2pt}|@{\hskip2pt} @{\hskip1pt}c@{\hskip1pt}|@{\hskip2pt}c@{\hskip1pt}|@{\hskip2pt}c@{\hskip1pt}|c@{\hskip1pt}|c@{\hskip1pt}|c@{\hskip1pt}|c@{\hskip1pt}|c@{\hskip1pt}|@{\hskip2pt}c@{\hskip1pt}}
        \toprule
        \multirow{2}{*}{\bf Model} & BoolQ &CB&COPA&MultiRC&ReCoRD&RTE&WiC&WSC& Average\\ 
         & Acc & F1/Acc & Acc &F1a/EM&F1/EM&Acc&Acc&Acc& Score\\
        \midrule
%%        DeBERTa$_{1.5B}$  &90.3  & 98.3/98.6 & 97 &89.1/66.3&93.9/93.7&94.5&76.8 &96.1&- \\ 
%%        DeBERTa$_{1.5B}$+SiFT  &90.9  & 98.6/98.8 & 98 &89.6/66.7&94.5/94.1&95.0&77.1 &96.6&- \\ 
        RoBERTa\textsubscript{large} & 87.1  & 90.5/95.2 & 90.6 &84.4/52.5&90.6/90.0&88.2&69.9 &89.0 &84.6\\  \hline
        NEXHA-Plus                  & 87.8  & 94.4/96.0 & 93.6 &84.6/55.1&90.1/89.6&89.1&74.6 &93.2 &86.7\\  \hline
        T5$_{11B}$ &91.2  & 93.9/96.8 & 94.8 &88.1/63.3&94.1/93.4&92.5&76.9 &93.8 &89.3\\ 
        \hline
        T5$_{11B}$+Meena &\textbf{91.3}  & \textbf{95.8/97.6} & 97.4 &88.3/63.0&94.2/93.5&92.7&\textbf{77.9} &95.9 &90.2\\ 
        \hline \hline
        Human  &89.0 & 95.8/98.9 & 100.0 &81.8/51.9&91.7/91.3&93.6&80.0 &100.0 &89.8\\  \hline \hline
        DeBERTa$_{1.5B}$+SiFT &90.4  & 94.9/97.2 & 96.8&\textbf{88.2/63.7}&\textbf{94.5/94.1}&\textbf{93.2}&76.4 &	\textbf{95.9} & 89.9\\  
        DeBERTa$_{Ensemble}$ &90.4  & 95.7/97.6 & \textbf{98.4}&88.2/63.7&94.5/94.1&93.2&77.5 &	95.9 & \textbf{90.3}\\
        \bottomrule
        \end{tabular}
    \caption{
    %Comparison results on the SuperGLUE(The top half of the table is the results on the development set and the bottom half of the table is the results on the testset.) 
    SuperGLUE test set results scored using the SuperGLUE evaluation server. All the results are obtained from \href{https://super.gluebenchmark.com}{https://super.gluebenchmark.com} on January 6, 2021.
    }
    \label{tab:superglue}
    
\end{table*}
%Based on the 1.5B pre-trained DeBERTa model, during the fine-tuning stage, we introduce a new adversarial training algorithm, called Scale-Invariant-Fine-Tuning(SiFT). The SiFT algorithm is derived from \citep{miyato2018vat,jiang2019smart}. Unlike existing adversarial training on NLP tasks which apply perturbation directly on the embedding space, SiFT adds a normalization layer right before of the perturbation layer , so thus the perturbation is applied on an uniform embedding space. Although the SiFT algorithm shows substantially gain over plain fine-tuning, it needs further study on more datasets to come to a  clear conclusion. We will leave it in our future exploration.
%To evaluate DeBERTa$_{1.5B}$ on downstream tasks, we run fine-tuning experiments on SuperGLUE benchmark \citep{roemmele2011choice,khashabi2018looking,zhang2018record,dagan2006pascal,bar2006second,giampiccolo2007third,bentivogli2009fifth,pilehvar2018wic,rudinger2018winogender,poliak2018dnc,levesque2011winograd}.

\section{Conclusions}
\label{sec:conclusion}

This paper presents a new model architecture DeBERTa (Decoding-enhanced BERT with disentangled attention) that improves the BERT and RoBERTa models using two novel techniques. 
The first is the disentangled attention mechanism, where
each word is represented using two vectors that encode its content and position, respectively, and the attention weights among words are computed using disentangled matrices on their contents and relative positions, respectively. 
The second is an enhanced mask decoder which incorporates absolute positions in the decoding layer to predict the masked tokens in model pre-training. 
In addition, a new virtual adversarial training method is used for fine-tuning to improve model's generalization on downstream tasks.

We show through a comprehensive empirical study that these techniques significantly improve the efficiency of model pre-training and the performance of downstream tasks. 
The DeBERTa model with 1.5 billion parameters surpasses the human
performance on the SuperGLUE benchmark for the first time
in terms of macro-average score.

DeBERTa surpassing human performance on SuperGLUE marks an important milestone toward general AI. Despite its promising results on SuperGLUE, the model is by no means reaching the human-level intelligence of NLU. Humans are extremely good at leveraging the knowledge learned from different tasks to solve a new task with no or little task-specific demonstration. This is referred to as \emph{compositional generalization}, the ability to generalize to novel compositions (new tasks) of familiar constituents (subtasks or basic problem-solving skills). Moving forward, it is worth exploring how to make DeBERTa incorporate compositional structures in a more explicit manner, which could allow combining neural and symbolic computation of natural language similar to what humans do.

\section{Acknowledgments}
%We thank Jade Huang and Nikos Karampatziakis from Microsoft Dynamics 365 AI for proofreading the draft with a lot of valuable advice, as well as Microsoft Research Technology Engineering team and Yoyo Liang from Microsoft Philly team for their help on the DGX-2 cluster. We thank Saksham Singhal, Xia Song, and Saurabh Tiwary from Microsoft Turing team for working with us on the 1.5B model with great support of their DGX-2 cluster. This effort is part of the Microsoft AI at Scale initiative\footnote{https://innovation.microsoft.com/en-us/ai-at-scale}.  We also thank the anonymous reviewers for valuable discussions.  

We thank Jade Huang and Nikos Karampatziakis for proofreading the paper and providing insightful comments. We thank Yoyo Liang, Saksham Singhal, Xia Song, and Saurabh Tiwary for their help with large-scale model training. 
%This effort is part of the Microsoft AI at Scale initiative\footnote{https://innovation.microsoft.com/en-us/ai-at-scale}.  
We also thank the anonymous reviewers for valuable discussions.  

%We would also like to thanks the  Microsoft Turing project for valuable discussions and comments and Microsoft Philly team and 

\bibliography{ref,ref_xiaodong}
\bibliographystyle{iclr2021_conference}
\clearpage
\appendix
\section{Appendix}
\label{sec:appendix}

\subsection{Dataset}
\begin{table*}[!htb]
	\begin{center}
		\begin{tabular}{l|l|c|c|c|c|c}
			\toprule 
			\bf Corpus &Task& \#Train & \#Dev & \#Test   & \#Label &Metrics\\ \hline \hline
			\multicolumn{6}{@{\hskip1pt}r@{\hskip1pt}}{\textbf{General Language Understanding Evaluation (GLUE})} \\ \hline
			%\multicolumn{6}{@{\hskip1pt}r@{\hskip1pt}}{Single-Sentence Classification} \\ \hline
			CoLA & Acceptability&8.5k & 1k & 1k & 2 & Matthews corr\\ \hline
			SST & Sentiment&67k & 872 & 1.8k & 2 & Accuracy\\ \hline
			%\multicolumn{6}{@{\hskip1pt}r@{\hskip1pt}}{Pairwise Text Classification} \\ \hline
			MNLI & NLI& 393k& 20k & 20k& 3 & Accuracy\\ \hline
            RTE & NLI &2.5k & 276 & 3k & 2 & Accuracy \\ \hline
            WNLI & NLI &634& 71& 146& 2 & Accuracy \\ \hline
			QQP & Paraphrase&364k & 40k & 391k& 2 & Accuracy/F1\\ \hline
            MRPC & Paraphrase &3.7k & 408 & 1.7k& 2&Accuracy/F1\\ \hline
			QNLI & QA/NLI& 108k &5.7k&5.7k&2& Accuracy\\ \hline 
		%	\multicolumn{5}{@{\hskip1pt}r@{\hskip1pt}}{Text Similarity} \\ \hline
			STS-B & Similarity &7k &1.5k& 1.4k &1 & Pearson/Spearman corr\\ \hline \hline
		%	\multicolumn{4}{@{\hskip1pt}r@{\hskip1pt}}{\textbf{Question Answering}} \\ \hline
			\multicolumn{6}{@{\hskip1pt}c@{\hskip1pt}}{\textbf{SuperGLUE}} \\ \hline
			WSC & Coreference& 554k & 104 & 146 & 2 & Accuracy\\ \hline
			%\multicolumn{6}{@{\hskip1pt}r@{\hskip1pt}}{Pairwise Text Classification} \\ \hline
			BoolQ & QA& 9,427 & 3,270 & 3,245 & 2 & Accuracy\\ \hline
            COPA & QA &400k & 100 & 500 & 2 & Accuracy \\ \hline
			CB & NLI& 250 & 57 & 250 & 3 & Accuracy/F1\\ \hline
            RTE & NLI &2.5k & 276 & 3k & 2 & Accuracy \\ \hline
            WiC & WSD &2.5k & 276 & 3k & 2 & Accuracy \\ \hline
			ReCoRD & MRC&  101k& 10k&10k &-& Exact Match (EM)/F1\\ \hline 

			%\multicolumn{4}{@{\hskip1pt}r@{\hskip1pt}}{Ranking} \\ \hline
			MultiRC & Multiple choice&  5,100& 953&1,800 &-& Exact Match (EM)/F1\\ \hline \hline
			\multicolumn{5}{@{\hskip1pt}r@{\hskip1pt}}{Question Answering} \\ \hline 
			SQuAD v1.1 & MRC&  87.6k& 10.5k&9.5k &-& Exact Match (EM)/F1\\ \hline 
			SQuAD v2.0 & MRC& 130.3k &11.9k& 8.9k&-& Exact Match (EM)/F1\\ \hline 			
			%\multicolumn{4}{@{\hskip1pt}r@{\hskip1pt}}{Ranking} \\ \hline
			%ReCoRD & MRC&  101k& 10k&10k &-& Exact Match (EM)/F1\\ \hline 
			RACE & MRC&  87,866& 4,887&4,934 &4& Accuracy\\ \hline
			SWAG & Multiple choice&  73.5k& 20k&20k &4& Accuracy\\ \hline \hline
			\multicolumn{4}{@{\hskip1pt}r@{\hskip1pt}}{\textbf{Token Classification}} \\ \hline 
			CoNLL 2003 & NER&  14,987& 3,466& 3,684 &8& F1\\
			
			\bottomrule

		\end{tabular}
	\end{center}
	\caption{Summary information of the NLP application benchmarks.
	}
	\label{tab:datasets}
\end{table*}
\noindent $\bullet$ \textbf{GLUE}. The General Language Understanding Evaluation (GLUE) benchmark is a collection of nine natural language understanding (NLU) tasks. As shown in Table~\ref{tab:datasets},
it includes question answering~\citep{squad1}, linguistic acceptability~\citep{cola2018}, sentiment analysis~\citep{sst2013}, text similarity~\citep{sts-b2017}, paraphrase detection~\citep{mrpc2005}, and natural language inference (NLI)~\citep{rte1,rte2,rte3,rte5,winograd2012,mnli2018}. The diversity of the tasks makes GLUE very suitable for evaluating the generalization and robustness of NLU models. 

%\noindent $\bullet$ \textbf{ReCoRD}  is a commonsense Question Answering dataset. Each example consists of a news article, drawn from CNN and DailyMail, and a Cloze-style question about the article in which one entity is masked out \citep{zhang2018record}.

\noindent $\bullet$ \textbf{SuperGLUE}. SuperGLUE is an extension of the GLUE benchmark, but more difficult, which is a collection of eight NLU tasks. It covers a various of tasks including question answering \citep{zhang2018record,clark2019boolq,khashabi2018multirc}, natural language inference \citep{rte1,rte2,rte3,rte5,de2019cb}, coreference resolution \citep{winograd2012} and word sense disambiguation \citep{pilehvar2019wic}.

\noindent $\bullet$ \textbf{RACE}  is a large-scale machine reading comprehension dataset, collected from English examinations in China, which are designed for middle school and high school students \citep{lai2017race}. 

\noindent $\bullet$ \textbf{SQuAD v1.1/v2.0} is the Stanford Question
Answering Dataset (SQuAD) v1.1 and v2.0 \citep{squad1,squad2} are popular machine reading comprehension benchmarks. Their passages come from approximately 500 Wikipedia articles and the questions and answers are obtained by crowdsourcing. The SQuAD v2.0 dataset includes unanswerable questions about the same paragraphs.

\noindent $\bullet$ \textbf{SWAG} is a large-scale adversarial dataset for the task of grounded commonsense inference, which unifies natural language inference and physically grounded reasoning \citep{zellers2018swag}. SWAG consists of 113k multiple choice questions about grounded situations.

\noindent $\bullet$ \textbf{CoNLL 2003}  is an English dataset consisting of text from a wide variety of sources. It has 4 types of named entity.

\subsection{Pre-training Dataset}
\label{appendix:pre-train}
For DeBERTa pre-training, we use Wikipedia (English Wikipedia dump\footnote{https://dumps.wikimedia.org/enwiki/}; 12GB), BookCorpus \citep{bookcorpus} \footnote{https://github.com/butsugiri/homemade\_bookcorpus} (6GB), OPENWEBTEXT (public Reddit content \citep{Gokaslan2019OpenWeb}; 38GB) and STORIES\footnote{https://github.com/tensorflow/models/tree/master/research/lm\_commonsense} (a subset of CommonCrawl \citep{trinh2018simple}; 31GB). The total data size after data deduplication\citep{shoeybi2019megatron} is about 78GB. For pre-training, we also sample 5\% training data as the validation set to monitor the training process. Table~\ref{tab:data-comp} compares datasets used in different pre-trained models.

\begin{table*}[htb!]
    \centering
    \begin{tabular}{@{\hskip1pt}l@{\hskip3pt}|@{\hskip3pt}c@{\hskip3pt}|@{\hskip3pt} c@{\hskip3pt}|@{\hskip3pt} c@{\hskip3pt}|@{\hskip3pt} c@{\hskip3pt}|@{\hskip3pt}c@{\hskip3pt}|@{\hskip3pt}c@{\hskip3pt}|@{\hskip3pt}c@{\hskip3pt}} 
    \toprule
        \bf Model&  Wiki+Book & OpenWebText & Stories & CC-News&Giga5&ClueWeb&Common Crawl  \\
                 & 16GB & 38GB & 31GB & 76GB & 16GB &19GB & 110GB         \\ \hline
       BERT& \checkmark & & & & &\\ \hline
       XLNet& \checkmark & & & &\checkmark &\checkmark&\checkmark\\ \hline
       RoBERTa& \checkmark & \checkmark& \checkmark& \checkmark& &\\ \hline
       DeBERTa&  \checkmark & \checkmark& \checkmark& & &\\ 
       DeBERTa$_{1.5B}$&  \checkmark & \checkmark& \checkmark& \checkmark & &\\ 
        \bottomrule
        \end{tabular}
    \caption{
    Comparison of the pre-training data.
    }
    \label{tab:data-comp}
\end{table*}

\subsection{Implementation Details}
\label{subsec:details}
%For the pre-training, we use a batch size of 2048. The learning rate is set as 2e-4, with the number of warm-up steps as 10000. 
Following RoBERTa \citep{liu2019roberta}, we  adopt dynamic data batching. We also include span masking~\citep{joshi2019spanbert} as an additional masking strategy with the span size up to three. 
%For a single complete training of DeBERTa$_{large}$, we use 6 DGX-2 machines with 96 V100 GPUs to train the model for 1M steps and it takes about 20 days. 
We list the detailed hyperparameters of pre-training in Table~\ref{tbl:hyper-pre-train}. For pre-training, we use Adam \citep{kingma2014adam} as the optimizer with weight decay \citep{loshchilov2018fixing}. For fine-tuning, even though we can get better and robust results with RAdam\citep{liu2019radam} on some tasks, e.g. CoLA, RTE and RACE, we use Adam\citep{kingma2014adam} as the optimizer for a fair comparison.
For fine-tuning, we train each task with a hyper-parameter search procedure, each run takes about 1-2 hours on a DGX-2 node. All the hyper-parameters are presented in Table~\ref{tbl:hyper-ft}. The model selection is based on the performance on the task-specific development sets. 

Our code is implemented based on Huggingface Transformers\footnote{https://github.com/huggingface/transformers}, FairSeq\footnote{https://github.com/pytorch/fairseq} and Megatron \citep{shoeybi2019megatron}\footnote{https://github.com/NVIDIA/Megatron-LM}.

%pick the best model according to the performance on the task-specific dev sets. 
\begin{table*}[htb!]
    \centering
    \begin{tabular}{@{\hskip3pt}l@{\hskip2pt}|@{\hskip2pt} c@{\hskip2pt}|@{\hskip2pt} c@{\hskip2pt}|@{\hskip2pt} c@{\hskip2pt}|@{\hskip2pt} c@{\hskip2pt}}
        \toprule
          Hyper-parameter & DeBERTa$_{1.5B}$& DeBERTa$_{large}$ &  DeBERTa$_{base}$ &DeBERTa$_{base-ablation}$\\
        \midrule
        Number of Layers& 48& 24 &12 &12\\
        Hidden size&1536 &1024 &768 &768\\
        FNN inner hidden size & 6144& 4096 & 3072& 3072 \\
        Attention Heads &24 & 16 & 12 & 12\\
        Attention Head size &64 & 64 & 64& 64 \\
        Dropout & 0.1& 0.1 & 0.1& 0.1 \\
        Warmup Steps & 10k & 10k & 10k & 10k\\
        Learning Rates & 1.5e-4& 2e-4& 2e-4& 1e-4 \\
        Batch Size & 2k& 2k & 2k & 256\\
        Weight Decay & 0.01& 0.01 & 0.01& 0.01 \\
        Max Steps & 1M& 1M & 1M & 1M\\
        Learning Rate Decay & Linear& Linear & Linear& Linear \\
        Adam $\epsilon$ & 1e-6& 1e-6 & 1e-6& 1e-6 \\
        Adam $\beta_1$ & 0.9& 0.9 & 0.9& 0.9 \\
        Adam $\beta_2$ & 0.999& 0.999 & 0.999& 0.999 \\
        Gradient Clipping & 1.0& 1.0 & 1.0& 1.0 \\ \hline
        Number of DGX-2 nodes & 16& 6 & 4 &1 \\
        Training Time & 30 days& 20 days & 10 days & 7 days \\
        \bottomrule
        \end{tabular}
    \caption{    Hyper-parameters for pre-training DeBERTa. }
    \label{tbl:hyper-pre-train}
\end{table*}

\begin{table*}[htb!]
   
    \centering
    \begin{tabular}{@{\hskip3pt}l@{\hskip2pt}|@{\hskip2pt} c@{\hskip2pt}|@{\hskip2pt} c@{\hskip2pt}|@{\hskip2pt} c@{\hskip2pt}}
        \toprule
          Hyper-parameter & DeBERTa$_{1.5B}$& DeBERTa$_{large}$ &  DeBERTa$_{base}$\\
        \midrule
        Dropout of task layer & \{0,0.15,0.3\}& \{0,0.1,0.15\} & \{0,0.1,0.15\} \\
        Warmup Steps & \{50,100,500,1000\}& \{50,100,500,1000\} & \{50,100,500,1000\} \\
        Learning Rates & \{1e-6, 3e-6, 5e-6\}& \{5e-6, 8e-6, 9e-6, 1e-5\} & \{1.5e-5,2e-5, 3e-5, 4e-5\} \\
        Batch Size & \{16,32,64\}& \{16,32,48,64\} & \{16,32,48,64\} \\
        Weight Decay & 0.01 & 0.01 \\
        Maximun Training Epochs & 10& 10 & 10 \\
        Learning Rate Decay & Linear & Linear & Linear \\
        Adam $\epsilon$ & 1e-6 & 1e-6 & 1e-6 \\
        Adam $\beta_1$ & 0.9 & 0.9 & 0.9 \\
        Adam $\beta_2$ & 0.999 & 0.999 & 0.999 \\
        Gradient Clipping & 1.0 & 1.0 & 1.0 \\
        \bottomrule
        \end{tabular}
    \caption{
    Hyper-parameters for fine-tuning DeBERTa on down-streaming tasks. 
    }
     \label{tbl:hyper-ft}
\end{table*}

\subsubsection{Pre-training Efficiency}
\label{subse:eff}
\begin{figure}[htb!]
\centering  
\subfloat[Results on MNLI development]{
\includegraphics[width=0.49\linewidth]{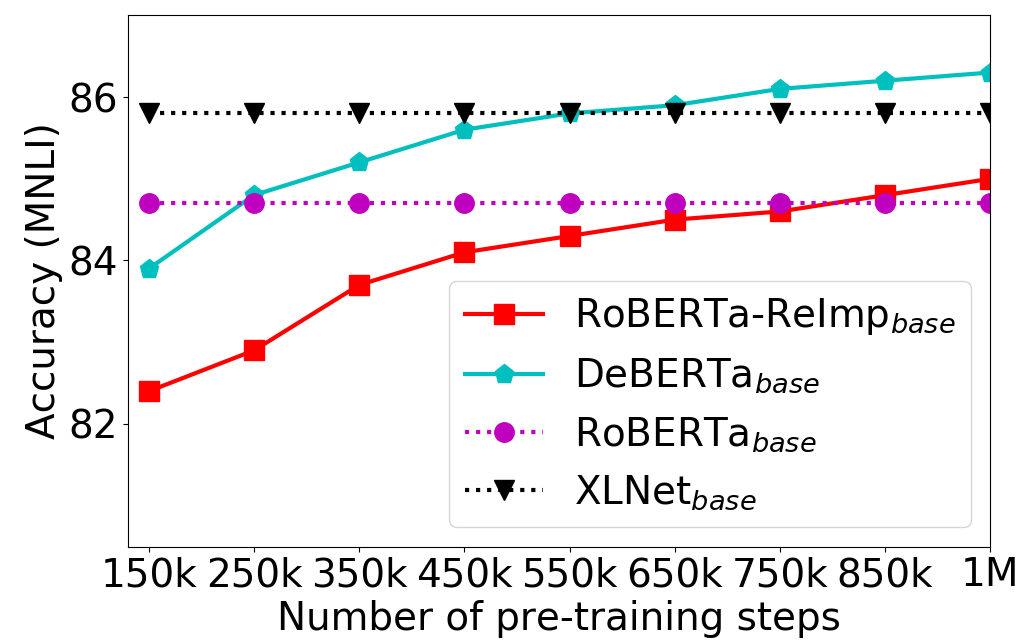}

}
\hfill
%\subfigure[Results on SQuAD v2.0 development]{\includegraphics[width=0.49\linewidth]{figures/deberta_squad_v2.png}}
\subfloat[Results on SQuAD v2.0 development]{\includegraphics[width=0.49\linewidth]{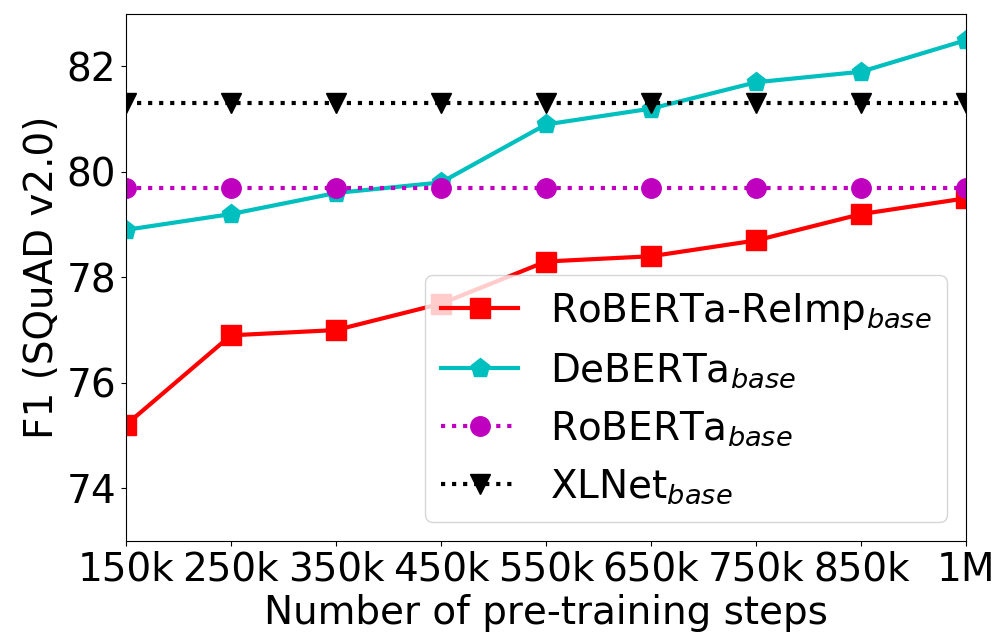}}
	\caption{Pre-training performance curve between {\ModelName} and its counterparts on the MNLI and SQuAD v2.0 development set.}
\label{fig:conv}
\end{figure}

To investigate the efficiency of model pre-training, we plot the performance of the fine-tuned model on downstream tasks 
as a function of the number of pre-training steps. 
As shown in Figure~\ref{fig:conv}, for RoBERTa-ReImp$_{base}$ and {\ModelName}$_{base}$, we dump a checkpoint every 150K pre-training steps, and then fine-tune the checkpoint on two representative downstream tasks, MNLI and SQuAD v2.0, and then report the accuracy and F1 score, respectively.  
As a reference, we also report the final model performance of both the original RoBERTa$_{base}$~\citep{liu2019roberta} and XLNet$_{base}$~\citep{yang2019xlnet}. 
% and plot them as flat dot lines.  
The results show that {\ModelName}$_{base}$ consistently outperforms RoBERTa-ReImp$_{base}$ during the course of pre-training.

\subsection{Main Results on Generation Tasks}
\label{subsec:generation}
In addition to NLU tasks, DeBERTa can also be extended to handle NLG tasks.
%to verify the impacts of the disentangled attention thoroughly in both settings. 
To allow DeBERTa operating like an auto-regressive model for text generation, we use a triangular matrix for self-attention and set the upper triangular part of the self-attention mask to $-\infty$, following~\cite{dong2019unilm}. 

We evaluate DeBERTa on the task of auto-regressive language model (ARLM) using Wikitext-103~\citep{merity2016pointer}. 
To do so, we train a new version of DeBERTa, denoted as DeBERTa-MT. 
It is jointly pre-trained using the MLM and ARLM tasks as in UniLM \citep{dong2019unilm}. 
The pre-training hyper-parameters follows that of DeBERTa$_{base}$ except that we use fewer training steps (200k). For comparison, we use RoBERTa as baseline, and include GPT-2 and Transformer-XL as additional references.  
DeBERTa-AP is a variant of DeBERTa where absolute position embeddings are incorporated in the input layer as RoBERTa. 
For a fair comparison, all these models are base models pre-trained in a similar setting.
%is the model with the same structure as RoBERTa base model except training with auto-regressive attention mask to enable left-to-right language model. Similarly, 
%DeBERTa$_{base}$ is the model with the same model sturcture as DeBERTa base model. 
%In addition, DeBERTa-MT$_{base}$ is the model similar to DeBERTa$_{base}$ except jointly training with MLM and ARLM. RoBERTa$_{base}$, DeBERTa$_{base}$ and DeBERTa-MT$_{base}$ all share the same training hyper-parameter as DeBERTa base model except for we only training with 200k steps. One difference between our model with Transformer-XL is that we use the vocabulary of GPT instead of build vocabulary on Wikitext-103 training data.

\begin{table*}[htb!]
    \centering
    \begin{tabular}{@{\hskip2pt}l|@{\hskip2pt} c @{\hskip2pt}|@{\hskip2pt} c @{\hskip2pt}| c@{\hskip2pt} |@{\hskip2pt}c@{\hskip2pt}|@{\hskip2pt} c@{\hskip2pt}|@{\hskip2pt}c @{\hskip2pt}}
        \toprule
        {\bf Model} &{RoBERTa}& {DeBERTa-AP} & {DeBERTa} & {DeBERTa-MT} &{GPT-2} &{Transformer-XL}  \\ 
        \midrule
        Dev PPL & 21.6 & 20.7 & 20.5 & \textbf{19.5} & -& 23.1  \\ \hline
        Test PPL & 21.6 & 20.0 & 19.9 & \textbf{19.5} &  37.50& 24  \\
        \bottomrule
        \end{tabular}
    \caption{
    Language model results in perplexity (lower is better) on Wikitext-103 . 
    }
    \label{tab:wiki}
 %   \vspace{-3mm}
\end{table*}
Table~\ref{tab:wiki} summarizes the results on Wikitext-103. 
We see that {\ModelName}\textsubscript{base} obtains lower perplexities on both dev and test data, and joint training using MLM and ARLM reduces perplexity further. %, showing the effectiveness of {\ModelName}. 
That DeBERTa-AP is inferior to DeBERTa indicates that it is more effective to incorporate absolute position embeddings of words in the decoding layer as the EMD in DeBERTa than in the input layer as RoBERTa.

\subsection{Handling long sequence input}
With relative position bias, we choose to truncate the maximum relative distance to $k$ as  in \eqref{dist}. Thus in each layer, each token can attend directly to at most $2(k-1)$ tokens and itself. By stacking Transformer layers, each token in the $l-$th layer can attend to at most $(2k-1)l$ tokens implicitly. Taking {\ModelName}$_{large}$ as an example, where $k=512, L=24$, in theory, the maximum sequence length that can be handled is 24,528. This is a byproduct benefit of our design choice and we find it beneficial for the RACE task. 
A comparison of long sequence effect on the RACE task is shown in Table~\ref{tab:race-long}.

\begin{table*}[htb!]
    \centering
    \begin{tabular}{@{\hskip2pt}l@{\hskip3pt}|@{\hskip2pt} c@{\hskip2pt}| @{\hskip2pt}c@{\hskip2pt}| @{\hskip2pt}c@{\hskip2pt}}
        \toprule
        Sequence length& Middle &High &Accuracy \\
        \midrule
        512 & 88.8 & 85.0 & 86.3 \\
        768 & 88.7 & 86.3 & 86.8 \\
        \bottomrule
    \end{tabular}
    \caption{
    The effect of handling long sequence input for RACE task with DeBERTa
    }
    \label{tab:race-long}
\end{table*}

Long sequence handling is an active research area. There have been a lot of studies where the Transformer architecture is extended for long sequence handling\citep{beltagy2020longformer, kitaev2020reformer, child2019generating, dai2019transformer}.  
One of our future research directions is to extend {\ModelName} to deal with extremely long sequences.
% and compare it with existing approaches.  
%While the capability of long sequence handling  of \ModelName \ needs further study on broader tasks, we believe \ModelName\ can be a good base for future research work on long sequence processing.

\subsection{Performance improvements of different model scales}
\label{subsec:scale}
In this subsection, we study the effect of different model sizes  applied to large models on GLUE. Table \ref{tab:xxlarge} summarizes the results,  showing that larger models can obtain a better result and SiFT also improves the model performance consistently.
%We progressively add each component on the top of the DeBERTa model. The results are summarized in Table~\ref{tab:xxlarge}. We observe that 1) sharing the projection matrix between different layers obtains a similar performance to the one without sharing, and reduces 14M parameters; 2) adding Convolutional neural network 
%To study the performance improvements of different models scales, in addition to the 1.5B model, we also train a 900M model with the same settings as the 1.5B model except that the 900M model only has 24 layers. We compare the model performance with GLUE benchmark. The results are shown in Table \ref{tab:xxlarge}.

\begin{table*}[htb!]
    \centering
    \begin{tabular}{@{\hskip3pt}l@{\hskip2pt}|@{\hskip2pt} c@{\hskip2pt}| @{\hskip2pt}c@{\hskip2pt}|c@{\hskip2pt}|c@{\hskip2pt}|c@{\hskip2pt}|c@{\hskip2pt}|c@{\hskip2pt}|c@{\hskip2pt}|c@{\hskip2pt}}
        \toprule
        \multirow{2}{*}{\bf Model} & {CoLA} &{QQP} &{MNLI-m/mm} &SST-2 &STS-B&QNLI&RTE&MRPC& Avg.\\ 
        & Mcc & Acc & Acc & Acc &Corr&Acc&Acc&Acc\\
        \midrule
        {\ModelName}$_{large}$ &70.5 & 92.3 & 91.1/91.1 & 96.8 & 92.8 &95.3&88.3& 91.9 &90.00\\
        {\ModelName}$_{900M}$ &71.1 & 92.3 & 91.7/91.6 & \textbf{97.5} & 92.0 & 95.8 &93.5& 93.1 &90.86\\
        {\ModelName}$_{1.5B}$ &72.0 & 92.7 & 91.7/91.9 & 97.2 & 92.9 &96.0 &93.9& 92.0 &91.17\\
        \hline
        {\ModelName}$_{1.5B}$+SiFT &\textbf{73.5} & \textbf{93.0} & \textbf{92.0/92.1} & 97.5 & \textbf{93.2} &\textbf{96.5} &\textbf{96.5}& \textbf{93.2} &\textbf{91.93}\\
        \bottomrule
        \end{tabular}
    \caption{
    Comparison results of DeBERTa models with different sizes on the GLUE development set. 
    }
    \label{tab:xxlarge}
\end{table*}

\begin{table*}[htb!]
    \centering
    \begin{tabular}{@{\hskip1pt}l|c| c| c| c}
    \toprule
        \bf Model&Parameters &{MNLI-m/mm}& {SQuAD v1.1} &{SQuAD v2.0} \\ 
          &\ &  Acc& F1/EM & F1/EM  \\ \hline
        
        RoBERTa-ReImp$_{base}$ & 120M & 84.9/85.1 & 91.1/84.8 &79.5/76.0  \\ 
        \hline
        DeBERTa$_{base}$ & 134M& 86.3/86.2 &92.1/86.1 & 82.5/79.3  \\
        + ShareProjection & 120M & 86.3/86.3 &92.2/86.2 & 82.3/79.5  \\
        + Conv & 122M & 86.3/86.5 &92.5/86.4 & 82.5/79.7  \\
        + 128k Vocab & 190M & 86.7/86.9 &93.1/86.8 & 83.0/80.1  \\
       \bottomrule
        \end{tabular}
    \caption{
    Ablation study of the additional modifications in DeBERTa$_{1.5B}$ and DeBERTa$_{900M}$ models. Note that we progressively add each component on the top of DeBERTa\textsubscript{base}. 
    }
    \label{tab:v2}
    %\vspace{-3mm}
\end{table*}

\subsection{Model complexity}
With the disentangled attention mechanism, we introduce three additional sets of parameters $\bm{W_{q,r}},\bm{W_{k,r}} \in R^{d \times d}$ and $\bm{P} \in R^{2k \times d}$. The total increase in model parameters is $2L\times d^2 + 2k\times d$. For the large model $(d=1024,L=24,k=512)$, this amounts to about $49M$ additional parameters, an increase of $13\%$. 
For the base model$(d=768,L=12,k=512)$, this amounts to $14M$ additional parameters, an increase of $12\%$. 
However, by sharing the projection matrix between content and position embedding, i.e.  $\bm{W_{q,r}}=\bm{W_{q,c}},\bm{W_{k,r}}=\bm{W_{k,c}}$, the number of parameters of DeBERTa is the same as RoBERTa. 
Our experiment on base model shows that the results are almost the same, as in Table~\ref{tab:v2}. 
%Due to limited computer resource, we didn't run this setting with large model and 

The additional computational complexity is $O(Nkd)$ due to the calculation of the additional \textit{position-to-content} and \textit{content-to-position} attention scores. 
Compared with BERT or RoBERTa, this increases the computational cost by $30\%$. 
Compared with XLNet which also uses relative position embedding, the increase of computational cost is about $15\%$. 
A further optimization by fusing the attention computation kernel can significantly reduce this additional cost. 
For $\DecoderName$, since the decoder in pre-training only reconstructs the masked tokens, it does not introduce additional computational cost for unmasked tokens. 
%used as the query in multi-step language model head, which is
In the situation where $15\%$ tokens are masked and we use only two decoder layers, 
the additional cost is $0.15 \times 2/L$ which results in an additional computational cost of only $3\%$ for base model($L=12$) and $2\%$ for large model($L=24$) in EMD. 
%The total increased computation cost of our model is acceptable and 

\subsection{Additional Details of Enhanced Mask Decoder}

\begin{figure}[]
\centering  
\subfloat[BERT decoding layer]{
    \includegraphics[trim={0.5mm 0.5mm 0.5mm 0.5mm},clip,width=0.35\linewidth]{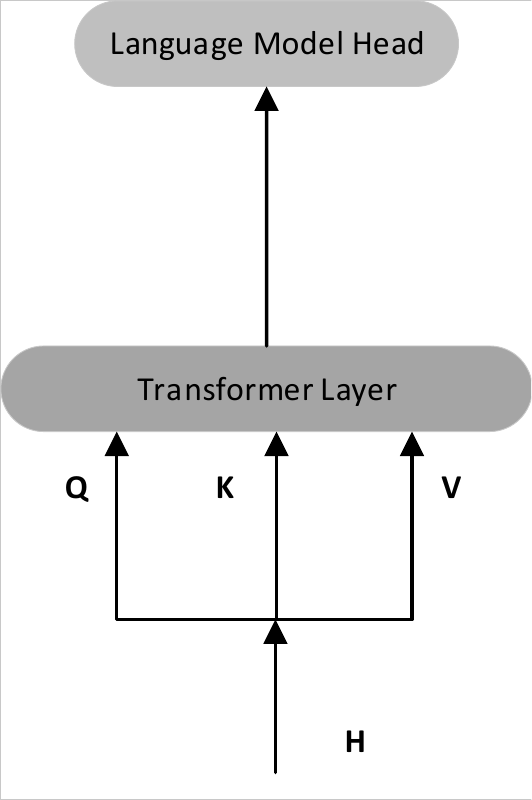}
    \label{fig:bert-a}
}
\hfill
\subfloat[Enhanced Mask Decoder]{
    \includegraphics[trim={0.5mm 0.5mm 0.5mm 0.5mm},clip,width=0.58\linewidth]{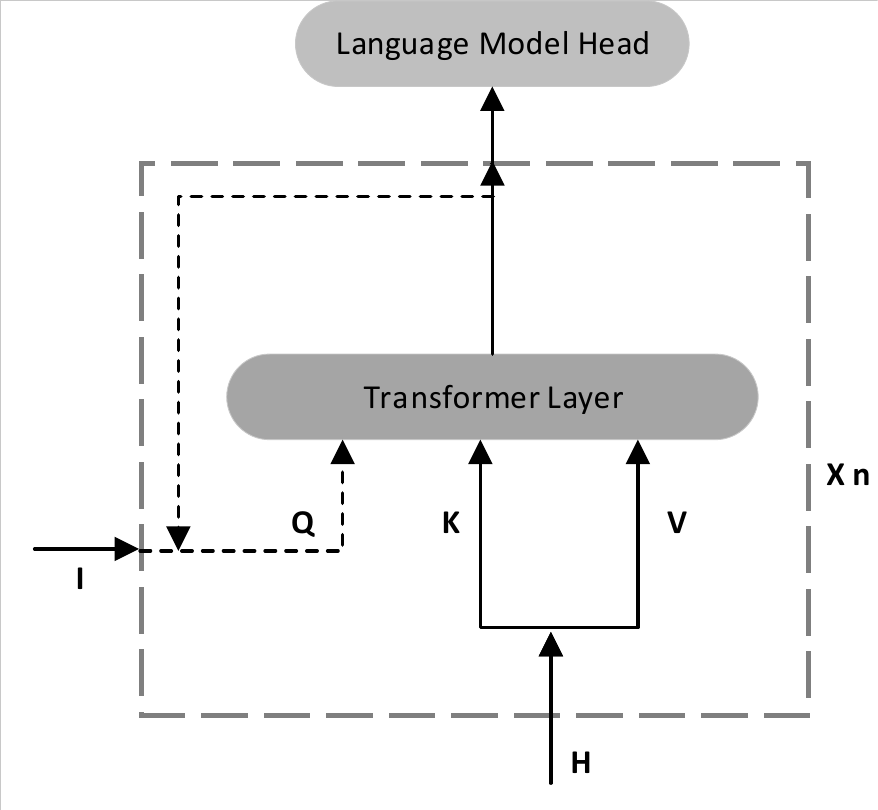}
    \label{fig:emd-b}
}

\caption{Comparison of the decoding layer.}
\label{fig:emd}
\end{figure}

The structure of EMD is shown in Figure~\ref{fig:emd-b}. There are two inputs for EMD, % $\DecoderName$, 
(i.e., $I, H$).  
$H$ denotes the hidden states from the previous Transformer layer, and $I$ can be any necessary information for decoding, e.g., $H$, absolute position embedding or output from previous EMD layer. 
$n$ denotes $n$ stacked layers of EMD where the output of each EMD layer will be the input $I$ for next EMD layer and the output of last EMD layer will be fed to the language model head directly. 
The $n$ layers can share the same weight. 
In our experiment we share the same weight for $n=2$ layers to reduce the number of parameters and use absolute position embedding as $I$ of the first EMD layer. 
When $I=H$ and $n=1$, EMD is the same as the BERT decoder layer. 
However, EMD is more general and flexible as it can take various types of input information for decoding.

\subsection{Attention Patterns}
\label{subsec:attention}
%\subsubsection{Attention Patterns}

\begin{figure}[htb!]
\centering 
\includegraphics[width=1\linewidth]{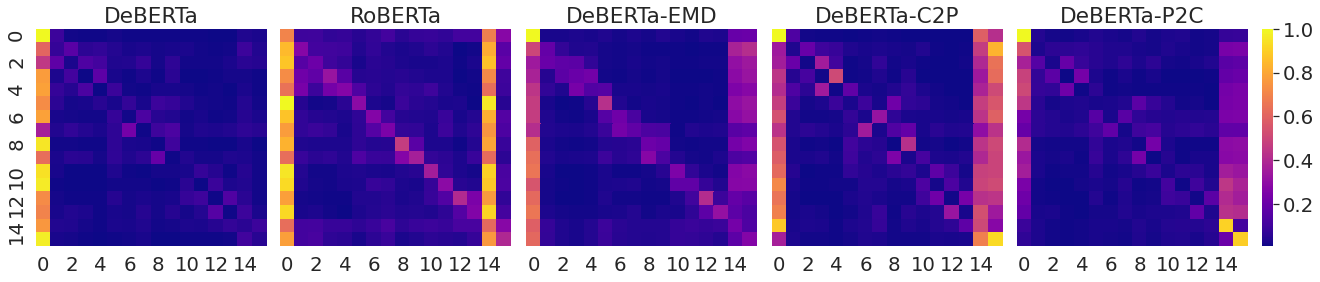}
\caption{Comparison  of attention patterns of the last layer among DeBERTa, RoBERTa and  DeBERTa variants (i.e., DeBERTa without EMD, C2P and P2C respectively). }
\label{fig:att-map}
\end{figure}

To visualize how DeBERTa operates differently from RoBERTa, we present in Figure~\ref{fig:att-map} the attention patterns (taken in the last self-attention layers) of RoBERTa, DeBERTa and three DeBERTa variants.
%in the last self-attention layer , where
%we also depict the attention patterns of the three DeBERTa variants %RoBERTa\textsubscript{base}, {\ModelName}\textsubscript{base} and its three variants 
%for comparison. 
%We also compare the attention patterns of {\ModelName} and RoBERTa.  As shown in 
%Comparing RoBERTa with {\ModelName}, 
We observe two differences. 
First, RoBERTa has a clear diagonal line effect for a token  attending to itself. But this effect is not very visible in {\ModelName}. 
%This can be attributed to the use of EMD, in which the vectors of the masked but unchanged tokens are replaced with their position embeddings,
This can be attributed to the use of EMD, in which the absolute position embedding is added to the hidden state of content as the query vector,
%rather than the original content embedding. 
% Another similar observation comes from 
as verified by the attention pattern of DeBERTa-EMD where the diagonal line effect is more visible  
%more brilliant 
than that of the original {\ModelName}. 
Second, we observe vertical strips in the attention patterns of RoBERTa, which are mainly caused by high-frequent functional words or tokens (e.g., ``a'', ``the'', and punctuation). 
For {\ModelName}, the strip only appears in the first column, which represents the \texttt{[CLS]} token. 
We conjecture that a dominant emphasis on \texttt{[CLS]} is desirable since the feature vector of \texttt{[CLS]} is often used as a contextual representation of the entire input sequence in downstream tasks. 
%On the other hand, if we look at the patterns the three RoBERTa variants, 
We also observe that the vertical strip effect is quite obvious in the patterns of the three {\ModelName} variants. 

\begin{figure}[htb!]
\centering 
\subfloat[]{
\includegraphics[width=0.65\linewidth]{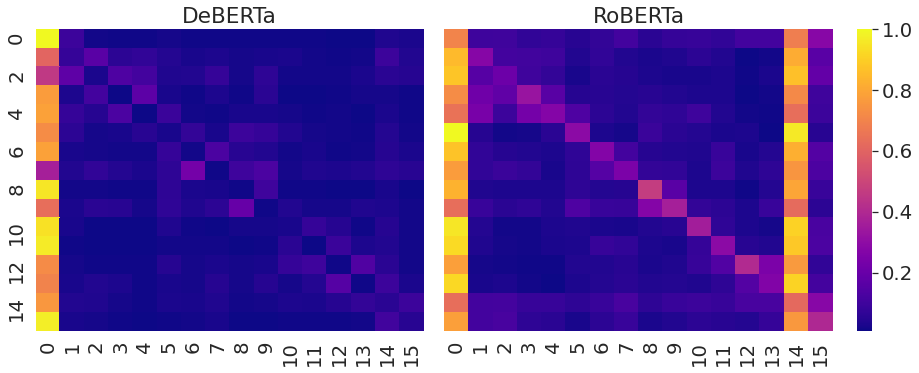}
}
\vfill
\subfloat[]{
\includegraphics[width=0.65\linewidth]{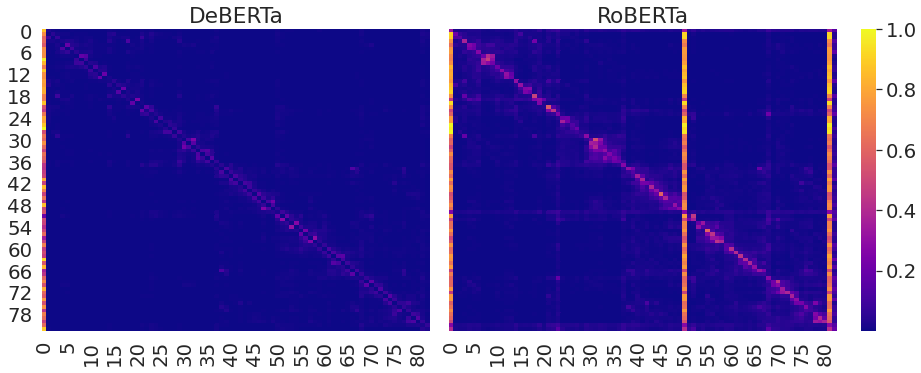}
}
\vfill
\subfloat[]{
\includegraphics[width=0.65\linewidth]{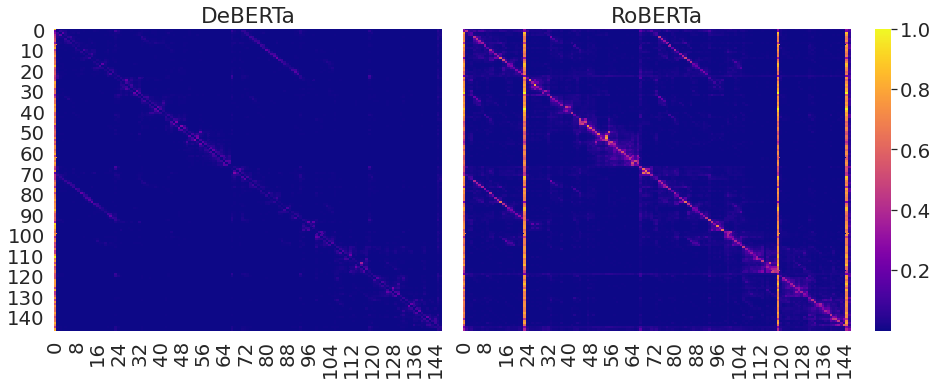}
}
\caption{Comparison  on attention patterns of the last layer between DeBERTa and RoBERTa.   }
\label{fig:att-01}
\end{figure}

\begin{figure}[htb!]
\centering 
\subfloat[]{
\includegraphics[width=0.98\linewidth]{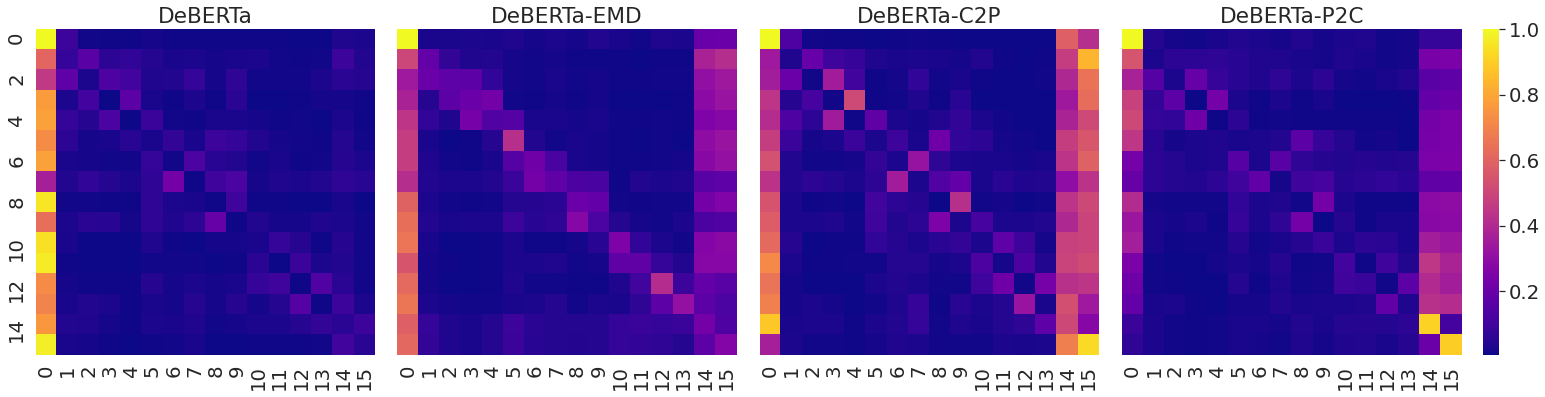}
\label{a1}
}
\vfill
\subfloat[]{
\includegraphics[width=0.98\linewidth]{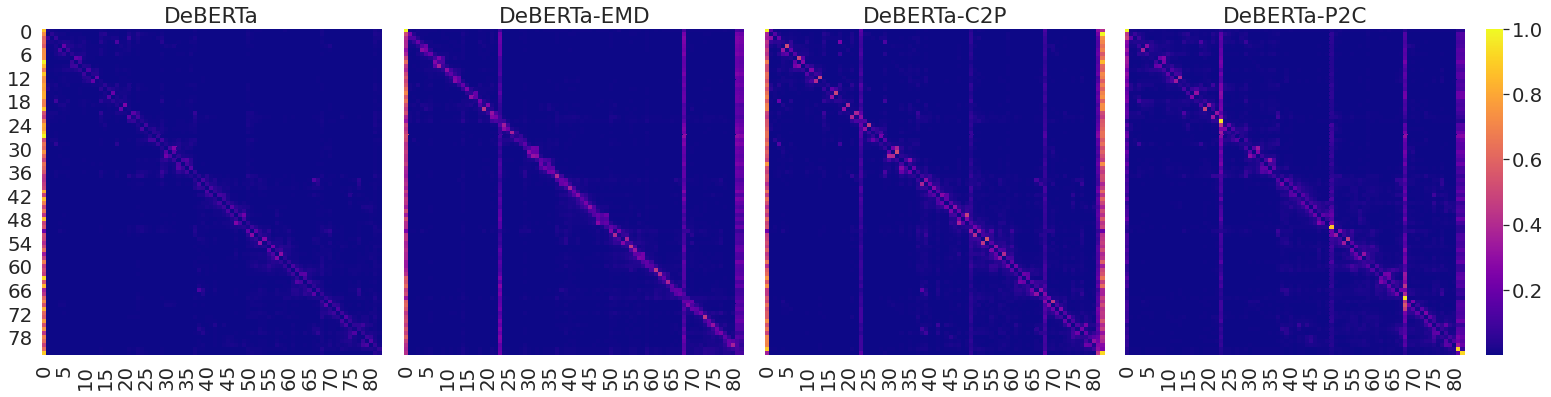}
\label{a2}
}
\vfill
\subfloat[]{
\includegraphics[width=0.98\linewidth]{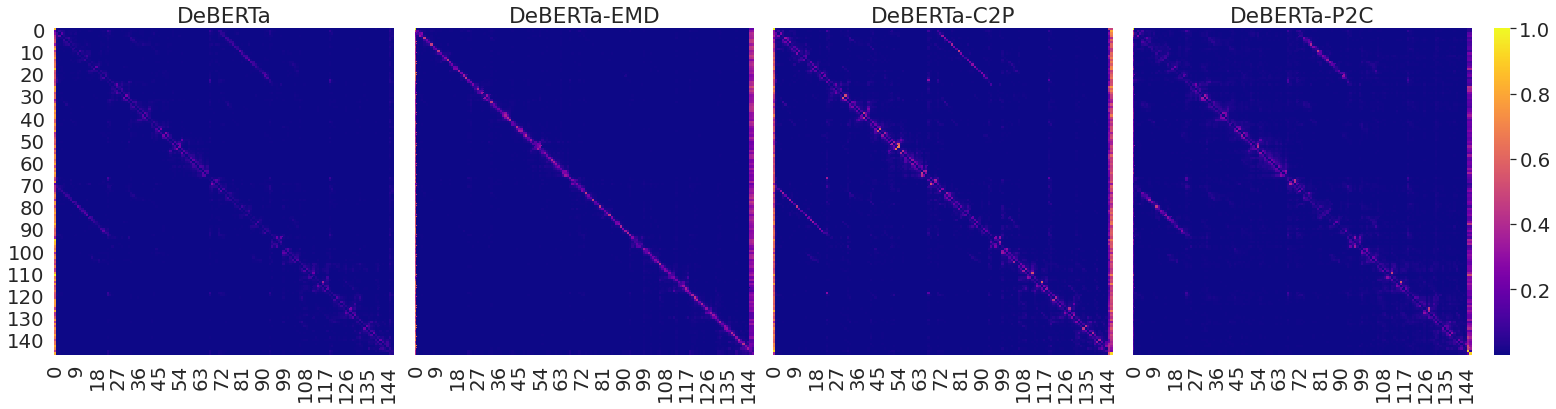}
\label{a3}
%\subcaption{$Q^{*}$}
}
\caption{Comparison  on attention patterns of last layer between DeBERTa and its variants (i.e. DeBERTa without EMD, C2P and P2C respectively).}
\label{fig:att}
\end{figure}

We present three additional examples to illustrate the different attention patterns of {\ModelName} and RoBERTa in Figures~\ref{fig:att-01} and~\ref{fig:att}.  

\clearpage

\subsection{Account for the Variance in Fine-Tuning}
Accounting for the variance of different runs of fine-tuning, in our experiments, we always follow ~\citep{liu2019roberta} to report the results on downstream tasks by averaging over five runs with different random initialization seeds, and perform significance test when comparing results.  
As the examples shown in Table~\ref{tab:var}, DeBERTa$_{base}$ significantly outperforms RoBERTa$_{base}$ ($p$-value < 0.05).

\begin{table}[ht]
    \centering
    \begin{tabular}{l | c | c | c}
    \toprule
    Model & MNLI-matched (Min/Max/Avg)&  SQuAD v1.1 (Min/Max/Avg) & $p$-value\\ \hline
    RoBERTa$_{base}$ &84.7/85.0/84.9 &90.8/91.3/91.1 & 0.02\\ \hline      
    DeBERTa$_{base}$ &86.1/86.5/86.3 &91.8/92.2/92.1 & 0.01\\
    \bottomrule
    \end{tabular}
    \caption{Comparison of DeBERTa and RoBERTa on MNLI-matched and SQuAD v1.1. }
    \label{tab:var}
\end{table}
% 	DeBERTa base(Min/max/avg)	RoBERTa-ReImp base(Min/max/avg)	p-value of t-tests
% MNLI-matched(Acc)	86.1/86.5/86.3	84.7/85.0/84.9	0.02
% SQUADv1.1(F1)	91.8/92.2/92.1	90.8/91.3/91.1	0.01

% \begin{figure}[htb!]
% \ContinuedFloat
% \centering 
% \setcounter{subfigure}{2}
% \subfigure[]{
% \includegraphics[width=0.95\linewidth]{figures/attention_map_07_1.png}
% \label{a3}
% %\subcaption{$Q^{*}$}
% }
% \caption{Comparison the attention map of last layer of DeBERTa with different components.}
% \label{fig:att}
% \end{figure}

%\xiaodl{How about: efficient DeBERTa as subsection?}
\subsection{Further improve the model efficiency}
In addition to scaling up transformer models with billions or trillions of parameters \citep{raffel2019t5,brown2020language,fedus2021switch}, it is important to improve model's parameter efficiency \citep{Kanakarajan2021SmallBenchNB}. In \ref{subse:eff} we have shown that DeBERTa is more parameter efficient than BERT and RoBERTa. 
In this section, we show further improvements in terms of parameter efficiency.

%\xiaodl{I'm not sure if we can call it RTD? Or we just keep it simple, e.g., ele short for electra. You can use this new term for the new paper.}
Replaced token detection (RTD) is a new pre-training objective introduced by ELECTRA \citep{clark2020electra}. It has been shown to be more effective than masked language model (MLM)  \citep{devlin2018bert, liu2019roberta}. 
In DeBERTa, we replace the MLM objective with the RTD objective, and denote the new variant as DeBERTa$_{RTD}$. 
We pre-train DeBERTa$_{RTD}$ using small, base and large settings with the same 160GB data as DeBERTa$_{1.5B}$. Following \citep{meng2021coco}, we set the width of the generator the same as that of the discriminator, but set its depth only half of the  discriminator's depth. 
Other hyper-parameters remain the same as DeBERTa$_{base}$ or DeBERTa$_{large}$. For the new member DeBERTa$_{RTD_{small}}$, it has 6 layers with the same width as DeBERTa$_{RTD_{base}}$. We evaluate our models on MNLI and SQuAD v2 datasets.
Table~\ref{tab:delectra} summarizes the results.
%and report the numbers on their development sets in table \ref{tab:delectra}. 
We observe that both DeBERTa$_{RTD_{base}}$ and DeBERTa$_{RTD_{large}}$ significantly outperform other models. For example, DeBERTa$_{RTD_{large}}$ obtains 0.9 absolute improvement over DeBERTa\textsubscript{Large} (the previous SoTA model) on MNLI and SQuAD v2.0, respectively.
%By replacing MLM with RTD, both DeBERTa$_{RTD_{base}}$ and DeBERTa$_{RTD_{large}}$ significantly outperform previous models with a similar structure by about 2 points on these tasks. 
It is worth noting that DeBERTa$_{RTD_{large}}$ is on-par with DeBERTa$_{1.5B}$ while has only 1/3 parameters of DeBERTa$_{1.5B}$.
%Simultaneously, 
Furthermore, DeBERTa$_{RTD_{small}}$ even outperforms BERT$_{large}$ by a large margin.
All these demonstrate the efficiency of DeBERTa$_{RTD}$ models and clearly show a huge potential to further improve model's parameter efficiency. Our work lays the base for future studies on far more parameter-efficient pre-trained language models. %\citep{Kanakarajan2021SmallBenchNB}.
\begin{table}[ht]
    \centering
    \begin{tabular}{l | c  | c}
    \toprule
    Model & MNLI(m/mm Acc)&  SQuAD v2.0 (F1/EM) \\ 
    \hline
    BERT$_{RTD_{small}}$ &88.2/87.9 & 82.9/80.4\\ \hline \hline 
    BERT$_{base}$ &84.3/84.7 & 76.3/73.7\\ 
    RoBERTa$_{base}$ &87.6/- & 83.7/80.5\\ 
    ELECTRA$_{base}$ &88.8/- &83.3/80.5\\ \hline 
    DeBERTa$_{base}$ &88.8/88.5 & 86.2/83.1 \\ 
    DeBERTa$_{RTD_{base}}$ &\textbf{90.6/90.8} & \textbf{88.4/85.4} \\ \hline \hline   
    BERT$_{large}$ &86.6/- &81.8/79.0 \\ 
    RoBERTa$_{large}$ &90.2/90.2 & 89.4/86.5 \\
    ELECTRA$_{large}$ &90.9/- & 90.6/88.0\\ \hline 
    DeBERTa$_{large}$ &91.1/91.1 & 90.7/88.0 \\ 
    DeBERTa$_{RTD_{large}}$ &\textbf{92.0/91.9} & \textbf{91.5/89.0} \\ \hline \hline 
    DeBERTa$_{1.5B}$ &91.7/91.9 & 92.2/89.7 \\
    \bottomrule
    \end{tabular}
    \caption{Comparison of different variants of DeBERTa models on MNLI and SQuAD 2.0.}
%    \caption{Further improve the efficiency of DeBERTa with RTD objective}
    \label{tab:delectra}
\end{table}

\iffalse
\begin{table}[ht]
    \centering
    \begin{tabular}{l | c | c | c}
    \toprule
    Model & MNLI(m/mm Acc)&  SQuAD v1.1 (F1/EM) & SQuAD v2.0 (F1/EM) \\ 
    \hline
    BERT$_{RTD_{small}}$ &88.2/87.9 &91.2/84.6 & 82.9/80.4\\ \hline \hline 
    BERT$_{base}$ &84.3/84.7 &88.5/81.0 & 76.3/73.7\\ 
    RoBERTa$_{base}$ &87.6/- &91.5/84.6 & 83.7/80.5\\ 
    ELECTRA$_{base}$ &88.8/- &90.8/84.5 &83.3/80.5\\ \hline 
    DeBERTa$_{base}$ &88.8/88.5 &93.1/87.2 & 86.2/83.1 \\ 
    DeBERTa$_{RTD_{base}}$ &\textbf{90.6/90.8} &\textbf{93.9/88.4} & \textbf{88.4/85.4} \\ \hline \hline   
    BERT$_{large}$ &86.6/- &90.9/84.1 &81.8/79.0 \\ 
    RoBERTa$_{large}$ &90.2/90.2 &94.6/88.9 & 89.4/86.5 \\
    ELECTRA$_{large}$ &90.9/- &94.9/89.7 & 90.6/88.0\\ \hline 
    DeBERTa$_{large}$ &91.1/91.1 &\textbf{95.5/90.1} & 90.7/88.0 \\ 
    DeBERTa$_{RTD_{large}}$ &\textbf{92.0/91.9} &95.2/89.7 & \textbf{91.5/89.0} \\ \hline \hline 
    DeBERTa$_{1.5B}$ &91.7/91.9 &96.1/91.4 & 92.2/89.7 \\
    \bottomrule
    \end{tabular}
    \caption{Further improve the efficiency of DeBERTa with RTD objective}
    \label{tab:delectra}
\end{table}
\fi

\end{document}